\documentclass[lettersize,journal]{IEEEtran}
\usepackage{amsmath,amsfonts}
\usepackage{amssymb}
\usepackage{mathtools}
\usepackage{amsthm}
\usepackage{array}
\usepackage[caption=false,font=footnotesize,labelfont=sf,textfont=sf]{subfig}
\usepackage{textcomp}
\usepackage{stfloats}
\usepackage{url}
\usepackage{verbatim}
\usepackage{graphicx}
\usepackage{cite}
\usepackage{hyperref}
\usepackage{cleveref}
\usepackage{booktabs}

\theoremstyle{plain}

\theoremstyle{definition}

\theoremstyle{remark}

\DeclareMathOperator*{\argmax}{argmax}

\DeclarePairedDelimiterX{\SquareBrackets}[1]{[}{]}{#1}
\DeclarePairedDelimiterX{\RoundBrackets}[1]{(}{)}{#1}
\DeclarePairedDelimiterX{\DivergenceBrackets}[2]{[}{]}{#1\;\delimsize\|\;#2}

\NewDocumentCommand{\pr}{ O{p} r() }{
  \def\prArg{#2}\patchcmd{\prArg}{|}{\mid}{}{}#1\RoundBrackets{\prArg}}
\NewDocumentCommand{\p}{ r() }{\pr[p](#1)}
\NewDocumentCommand{\q}{ r() }{\pr[q](#1)}
\NewDocumentCommand{\prm}{ r() }{\pr[\mathrm{p}](#1)}
\NewDocumentCommand{\Normal}{ r() }{\pr[\operatorname{Normal}](#1)}
\NewDocumentCommand{\Cat}{ r() }{\pr[\operatorname{Cat}](#1)}
\NewDocumentCommand{\Beta}{ r() }{\pr[\operatorname{Beta}](#1)}
\NewDocumentCommand{\Bernoulli}{ r() }{\pr[\operatorname{Bernoulli}](#1)}
\NewDocumentCommand{\Dir}{ r() }{\pr[\operatorname{Dir}](#1)}

\newcommand{\E}[3][]{\mathbb{\mathbb{E}}_{#2}#1[#3#1]}

\newcommand{\KL}{\mathrm{\operatorname{KL}}\DivergenceBrackets}

\crefname{algocf}{Algorithm}{Algorithms}

\usepackage[disable,textsize=tiny]{todonotes}
\usepackage{threeparttablex}
\usepackage[linesnumbered,ruled,noend]{algorithm2e}
\usepackage{algpseudocode}
\usepackage{multirow}
\usepackage{colortbl}

\begin{document}

\title{Dream to Recall: Imagination-Guided Experience Retrieval for Memory-Persistent Vision-and-Language Navigation}

\author{Yunzhe Xu,
        Yiyuan Pan,
        Zhe Liu
\thanks{This paper was supported by the National Natural Science Foundation of China under Grant 62303307, and in part by the National Key Laboratory of Human Machine Hybrid Augmented Intelligence, Xi’an Jiaotong University (No. HMHAI-202408). (Corresponding author: Zhe Liu.)}
\thanks{The authors are with the School of Automation and Intelligent Sensing, Shanghai Jiao Tong University, Shanghai 200240, China (e-mail: xyz9911@sjtu.edu.cn; pyy030406@sjtu.edu.cn; liuzhesjtu@sjtu.edu.cn).}
\thanks{Code is available at \href{https://github.com/xyz9911/Memoir}{https://github.com/xyz9911/Memoir}.}
}



\maketitle

\begin{abstract}
Vision-and-Language Navigation (VLN) requires agents to follow natural language instructions through environments, with memory-persistent variants demanding progressive improvement through accumulated experience. Existing approaches for memory-persistent VLN face critical limitations: they lack effective memory access mechanisms, instead relying on entire memory incorporation or fixed-horizon lookup, and predominantly store only environmental observations while neglecting navigation behavioral patterns that encode valuable decision-making strategies. We present Memoir, which employs imagination as a retrieval mechanism grounded by explicit memory: a world model imagines future navigation states as queries to selectively retrieve relevant environmental observations and behavioral histories. The approach comprises: 1) a language-conditioned world model that imagines future states serving dual purposes: encoding experiences for storage and generating retrieval queries; 2) Hybrid Viewpoint-Level Memory that anchors both observations and behavioral patterns to viewpoints, enabling hybrid retrieval; and 3) an experience-augmented navigation model that integrates retrieved knowledge through specialized encoders. Extensive evaluation across diverse memory-persistent VLN benchmarks with 10 distinct testing scenarios demonstrates Memoir's effectiveness: significant improvements across all scenarios, with 5.4\% SPL gains on IR2R over the best memory-persistent baseline, accompanied by 8.3× training speedup and 74\% inference memory reduction. The results validate that predictive retrieval of both environmental and behavioral memories enables more effective navigation, with analysis indicating substantial headroom (73.3\% vs 93.4\% upper bound) for this imagination-guided paradigm.
\end{abstract}


\section{Introduction}
\IEEEPARstart{V}{ision}-and-Language Navigation (VLN) \cite{2018r2r} represents a cornerstone challenge in embodied AI, requiring agents to interpret natural language instructions and navigate through environments to reach specified goals. The fundamental episodic nature of traditional VLN tasks \cite{2018r2r, 2020reverie, 2020rxr}, where agents operate independently across episodes without retaining experiential knowledge, limits their capacity for progressive improvement and environmental adaptation, constraining real-world applicability where sustained operation is essential. This limitation has motivated the development of memory-persistent navigation tasks \cite{2020multion, 2023ivln, 2025gsa} that evaluate agents' ability to accumulate and leverage experience across multiple navigation episodes. These tasks more accurately reflect practical application scenarios where robotic agents must continuously improve their navigation capabilities through environmental familiarity and learned behavioral patterns.

\begin{figure*}[t]
\centering
\includegraphics[scale=0.56]{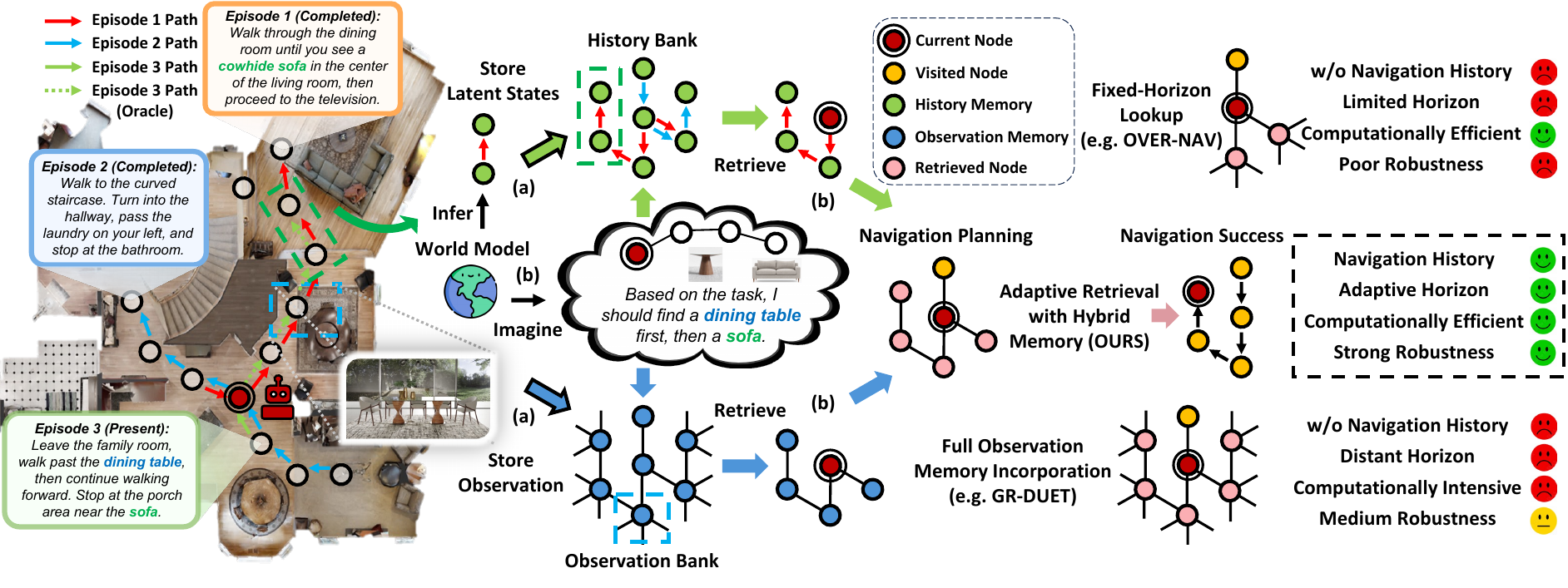}
\caption{Overview of Memoir's workflow for experience retrieval via imagination. \textbf{(a)} In previous episodes (1 and 2), the agent populates the history bank with latent states encoded by the world model, and fills the observation bank with observations. \textbf{(b)} In the current episode (3), the agent utilizes world model imagination to generate retrieval queries and retrieves memory from both memory banks at each viewpoint for navigation planning. Compared with GR-DUET \cite{2025gsa} that incorporates all retained observation memory and OVER-NAV \cite{2024overnav} that only applies fixed-horizon lookup, our approach adaptively retrieves both observation and histories for navigation planning through imagination.
}
\label{fig:intro}
\end{figure*}

Recent advances in memory-persistent VLN have primarily focused on long-term memory mechanisms for progressive scene knowledge accumulation. Early approaches employed strategies such as episodic history stacking \cite{2023ivln}, but simply extending history suffers from redundancy-induced performance degradation. Subsequent work \cite{2024esceme} addressed this by augmenting visual representations with broader spatial horizons rather than incorporating navigation histories, while OVER-NAV \cite{2024overnav} leverages open-vocabulary detection to construct multimodal topological graphs that strengthen keyword-observation correspondence. Most recently, GR-DUET \cite{2025gsa} enhanced the DUET architecture \cite{2022duet} with retained topological observation memory, achieving strong performance in VLN scene adaptation.

Despite these advances, existing approaches exhibit two critical limitations. \textbf{First}, current approaches lack effective memory access mechanisms, instead relying on either complete memory incorporation (leading to irrelevant information integration and computational overhead) or fixed-horizon spatial lookup (risking valuable experience loss). \textbf{Second}, navigation behavioral histories contain valuable decision-making patterns regarding how agents interpreted instructions and selected actions across different scenarios. However, existing memory-persistent VLN methods either ignore them entirely or, when attempted \cite{2023ivln}, fail to effectively leverage this information.


How can agents effectively determine which memories to access in order to leverage navigation experiences? Human navigators naturally engage in mental imagination of navigation routes \cite{2025humanneuraldynamics} and future travel events \cite{2021affectiveforecasting}, consulting experiences to finalize decisions based on mental simulations \cite{2025inspiringtourists}, highlighting that imagination serves as a query mechanism—agents can predict where they might navigate and retrieve relevant past experiences matching those predicted states. This paradigm differs from traditional imagine-planning approaches \cite{2023dreamwalker} that generate trajectories in isolation; instead, imagination is grounded by querying explicit long-term memory, ensuring retrieved experiences directly inform decision-making while avoiding hallucination. To this end, we propose \textbf{M}odel-based Hybrid Vi\textbf{e}wpoint-Level \textbf{M}em\textbf{o}ry for Exper\textbf{i}ence \textbf{R}etrieval (Memoir), an agent that employs predictive world modeling for memory retrieval at viewpoint granularity. Our approach addresses the aforementioned limitations through a unified framework. First, adaptive retrieval is grounded by using imagined future states as queries to selectively access verified experiences, avoiding both complete memory incorporation and fixed-horizon lookup. Second, behavioral pattern preservation is enabled by encoding navigation histories into latent states that capture decision-making strategies with viewpoint-level anchoring. \Cref{fig:intro} illustrates Memoir's workflow from memory storage to retrieval.

Realizing this imagination-guided paradigm requires addressing three challenges: how to generate predictive queries, what to store and retrieve, and how to integrate retrieved knowledge for navigation. Memoir tackles these through a unified framework. \textbf{1)} A language-conditioned world model learns to imagine future navigation states conditioned on instructions. These imagined states serve dual purposes: encoding current experience into latent representations for storage, and generating queries to retrieve similar past experiences. \textbf{2)} Hybrid Viewpoint-Level Memory (HVM) maintains this accumulated knowledge by anchoring both environmental observations and behavioral patterns to viewpoints, enabling retrieval of not just what agents saw, but how they navigated. \textbf{3)} The navigation model then processes current observations alongside retrieved experiences through specialized encoders to make informed decisions. This enables adaptive memory access that preserves strategic knowledge across episodes.

We implement Memoir on various VLN methods, validating its effectiveness across memory-persistent benchmarks with 10 distinctive testing scenarios. Memoir demonstrates consistent improvements, achieving 5.4\% improvement in SPL on IR2R \cite{2023ivln}, accompanied by 8.3× training speedup and 74\% inference memory reduction. We also reveal substantial headroom (73.3\% vs 93.4\% upper bound) for this paradigm and illuminate future directions. Our key contributions include:

\begin{itemize}
\item \textbf{A Novel Paradigm of Imagination-Guided Retrieval:} We propose a paradigm shift from passive memory accumulation to active, predictive retrieval. Unlike traditional methods that rely on full incorporation or fixed-horizon lookup, we utilize a language-conditioned world model to imagine future states as dynamic queries. This approach grounds imagination in verified experience, enabling the adaptive filtering of both environmental observations and behavioral histories based on navigation intent.

\item \textbf{Hybrid Viewpoint-Level Memory Architecture:} We introduce a unified memory architecture that anchors both environmental observations and behavioral decision patterns encoded by the world model to specific viewpoints. This allows the agent to leverage historical navigation strategies alongside visual context, a dimension largely neglected in prior DUET-style architectures.

\item \textbf{Efficient and Robust Navigation System:} We develop Memoir, which integrates specific VLN architectural innovations including a Navigation-History Encoder. Extensive evaluations demonstrate consistent improvements with significant efficiency benefits, while oracle analysis reveals substantial headroom, illuminating promising directions for advancing this paradigm.
\end{itemize}

\section{Related Work}

\subsection{Vision-and-Language Navigation}
Vision-and-Language Navigation (VLN) \cite{2018r2r,2020rxr,2020reverie} requires agents to follow natural language instructions while navigating toward target destinations. Single-episode VLN research has evolved through data augmentation approaches from speaker models \cite{2018speaker,2020prevalent,2023scalevln} to synthetic data \cite{2024navid, 2025flame} using Large Language Models (LLMs), and memory architectures progressing from historical representations \cite{2021recbert, 2021hamt} to structured spatial systems, particularly topological observation memory \cite{2021structuredscenedmemory,2022duet} which has been widely adopted. However, these single-episode approaches cannot accumulate knowledge across episodes. Memory-persistent VLN benchmarks \cite{2023ivln, 2025gsa} address real-world requirements where agents should operate continuously and improve through accumulated experience. TourHAMT \cite{2023ivln} extends historical memory by stacking complete navigation sequences, but suffers performance degradation from excessive redundancy. ESceme \cite{2024esceme} enhances environmental observations with broader spatial contexts, while OVER-NAV \cite{2024overnav} constructs omni-graphs with fixed-distance retrieval. MAP-CMA \cite{2023ivln} builds global semantic maps augmented with fixed-horizon egocentric perception. GR-DUET \cite{2025gsa} retains complete topological memory, achieving performance gains at computational cost. These approaches share fundamental limitations: reliance on complete memory incorporation or fixed-horizon lookup, and exclusive focus on environmental observations while neglecting navigation behavioral patterns that encode decision-making strategies across contexts. Our work addresses these limitations through imagination-guided memory retrieval that selectively accesses both environmental and behavioral histories.

\subsection{Memory Mechanism}
Memory mechanisms in navigation systems encompass two primary types that serve complementary roles in spatial reasoning. Navigation history memory captures temporal decision-making patterns and behavioral context through sequence representations \cite{2018neuralmap, 2021recbert} or natural language expression \cite{2025sevln, 2026matchnav}, preserving how agents make decisions across different scenarios. Environmental observation memory preserves spatial information through structured representations such as occupancy maps \cite{2018mapnet, 2020occupancyanticipation}, semantic maps \cite{2021semanticmapnet, 2023vlmaps}, bird's-eye view representations \cite{2024360bev, 2023bevscenegraph}, and topological memory \cite{2020topologicalslam, 2023topologicalsemanticgraph, 2024hierarchicalscenegraph}, maintaining spatial layouts and visual features for scene understanding. In memory-persistent scenarios where agents must accumulate knowledge across diverse experiences, current approaches \cite{2023ivln, 2025gsa} treat these information sources separately. Environmental observations alone cannot encode the behavioral reasoning underlying navigation decisions, while navigation histories without spatial anchoring cannot disambiguate similar patterns across different environments. This separation limits knowledge transfer across navigation scenarios, motivating our unified memory that leverages both spatial and temporal historical information.

\subsection{World Model}
Predictive world models have demonstrated significant impact in reinforcement learning through POMDP solutions via latent dynamics modeling \cite{2019rssm, 2020dreamtocontrol}. The Recurrent State-Space Model (RSSM) \cite{2019rssm} represents the dominant architecture, with extensions to language conditioning \cite{2024dynalang} and large-scale pretraining \cite{2024contextwm}. Contrastive world models \cite{2021celbo, 2021dreaming} offer computational efficiency without observation reconstruction. In navigation domains, world models \cite{2025navigationworldmodel} serve diverse purposes: future observation synthesis for data augmentation \cite{2023panogen, 2025navmorph}, trajectory planning through imagination \cite{2021pathdreamer, 2022crossmodalmap, 2023dreamwalker, 2025sali}, and auxiliary task formulation \cite{2023vlnfutureviewimagesemantics, 2025monodream}. While effective, using world models as surrogate environments often suffer from compounding hallucination errors in complex tasks. A nascent alternative is using world models for memory access. MBEC \cite{2021mbec} pioneered this direction in episodic control, but is limited to querying episodic buffers for policy optimization during training. Our approach advances this to a ``Dream to Recall" paradigm by unifying world modeling with retrieval for navigation reasoning. This repurposes the world model from a simulator to a neural search engine, anchoring imagination to grounded, long-term navigation experience.

\section{Preliminaries}

\subsection{VLN Formulation}

Vision-and-Language Navigation (VLN) requires an agent to follow instructions and navigate towards a target. The environment is represented as a connectivity graph $\mathcal{G} = (\mathcal{V}, \mathcal{E})$, where $\mathcal{V}$ denotes navigable viewpoints and $\mathcal{E}$ represents traversable edges connecting adjacent viewpoints.

\noindent\textbf{Single-Episode VLN Formulation.} In the traditional episodic setting, an agent receives a natural language instruction $\ell$ and is initialized at a starting viewpoint $v_1 \in \mathcal{V}$. At each timestep $t$, the agent observes a panoramic observation $o_t = \{o_t^{(i)}\}_{i=1}^{36}$ comprising 36 directional views: 12 horizontal viewing angles, each captured at three elevation levels (upward, horizontal, downward). The agent's action space at viewpoint $v_t$ includes navigation to any neighboring viewpoint $v_j \in \mathcal{N}(v_t)$ and a terminal stop action, where $\mathcal{N}(v_t) = \{v_j \in \mathcal{V} : (v_t, v_j) \in \mathcal{E}\}$ denotes the set of adjacent viewpoints. The episode terminates when the agent executes a stop action or reaches a maximum step limit $T_{\max}$.

\noindent\textbf{Memory-Persistent VLN Formulation.} While traditional VLN effectively evaluates basic instruction-following capabilities, it fails to capture the requirements of progressive improvement during persistent operation. Memory-persistent VLN addresses this limitation by introducing a persistent memory bank $\mathcal{M} = \{(\ell^{(k)}, \mathcal{G}^{(k)}, \mathcal{O}^{(k)}, \mathcal{A}^{(k)})\}_{k=1}^{N}$ that accumulates experiential knowledge across multiple episodes, where for $k$-th episode, $\ell^{(k)}$ is the instruction, $\mathcal{G}^{(k)} = (\mathcal{V}^{(k)}, \mathcal{E}^{(k)})$ is the observed subgraph after $k$ episodes, $\mathcal{O}^{(k)} = \{o_t^{(k)}\}_{t=1}^{T^{(k)}}$ and $\mathcal{A}^{(k)} = \{a_t^{(k)}\}_{t=1}^{T^{(k)}}$ records observations and actions respectively. The bank $\mathcal{M}$ is incrementally updated in each episode and serves as a persistent repository for decisions, enabling progressive performance improvement through accumulated environmental familiarity and learned behavioral patterns.

\begin{figure*}[t]
  \centering
  \includegraphics[scale=0.5]{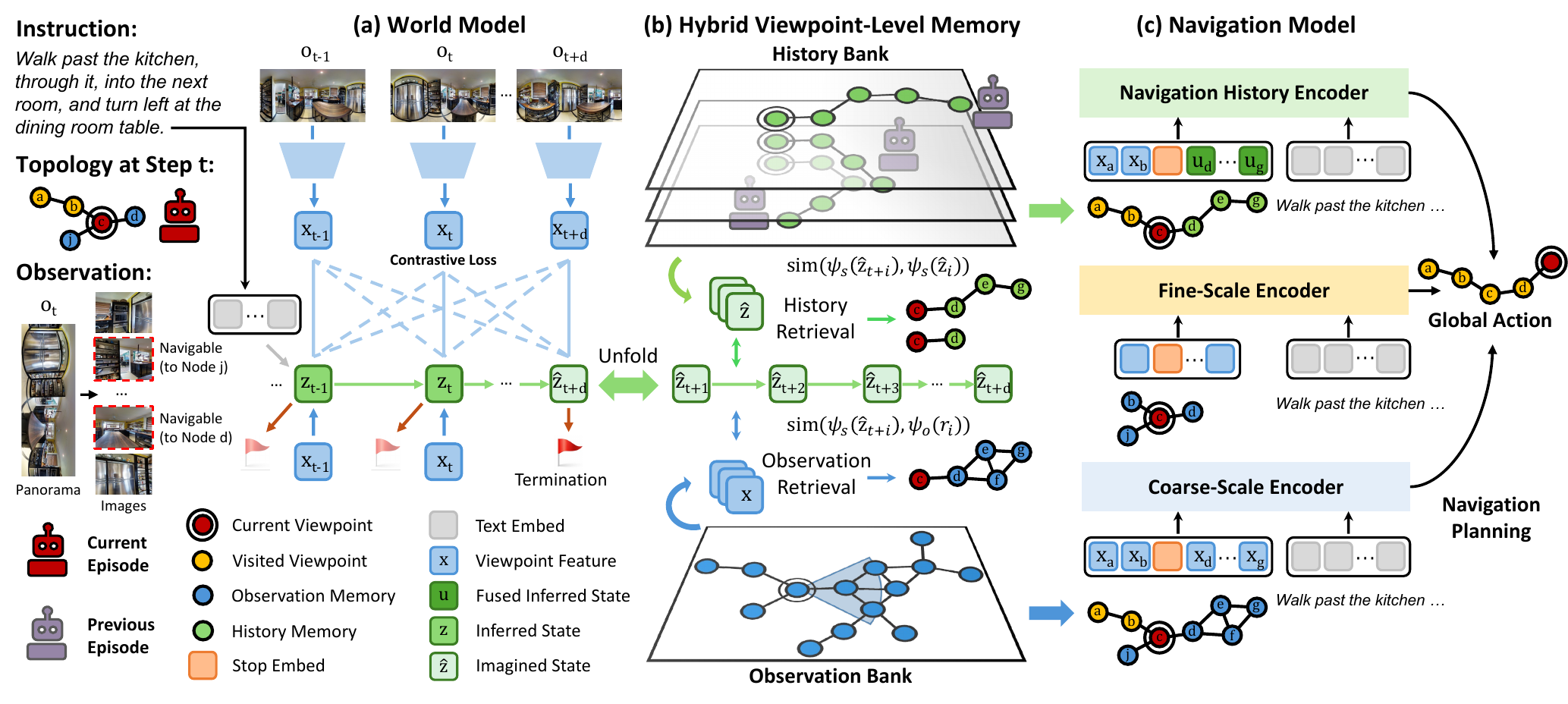}
  \caption{Details of imagination-guided experience retrieval. \textbf{(a)} The world model learns state-observation compatibility through contrastive training (top). During navigation, it infers the current state from observations and instruction, then recursively imagines future states (bottom). \textbf{(b)} Imagined trajectories enable dual retrieval: histories via state sequence similarity matching, and observations via topological searching based on state-observation compatibility. \textbf{(c)} Three specialized encoders process retrieved navigation histories, local observations, and retrieved observations respectively to determine the final action.}
  \label{fig:method}
\end{figure*}

\subsection{Dual-Scale Graph Transformer (DUET)}

DUET \cite{2022duet} enables topological navigation through topological mapping and global action planning.

\noindent\textbf{Topological Mapping.} The agent maintains an incrementally constructed topological representation $\mathcal{G}_t = (\mathcal{V}_t, \mathcal{E}_t)$ of the explored environment, where $\mathcal{G}_t \subseteq \mathcal{G}$ represents the observed subset after $t$ navigation steps. The viewpoint set $\mathcal{V}_t$ is partitioned into three categories: visited viewpoints, frontier viewpoints (observable but unvisited neighbors), and the current viewpoint. At each timestep $t$, the topological graph is updated by incorporating the current viewpoint $v_t$ and its navigable neighbors $\mathcal{N}(v_t)$ into $\mathcal{V}_{t-1}$, with corresponding edge updates to $\mathcal{E}_{t-1}$. Visual representations $r_t = \{r_t^{(i)}\}_{i=1}^{36}$ are computed through an observation encoder applied to $o_t$. The visual representation of the current viewpoint $x_t$ is obtained via average pooling of $r_t$, while each unvisited neighboring viewpoint $v_j \in \mathcal{N}(v_t)$ is represented by its corresponding directional embedding $r_t^{(i_j)}$ where $i_j$ denotes the view index oriented toward $v_j$. For viewpoints observed from multiple locations, embeddings are averaged to maintain consistency.

\noindent\textbf{Global Action Planning.} DUET combines coarse-scale planning over the topological graph with fine-scale planning over immediate neighbors. The instruction $\ell$ is processed through a transformer to obtain textual representations $\hat{\ell}$. For coarse-scale planning, node representations $x_j$ for viewpoints $v_j \in \mathcal{V}_t$ are augmented with a special stop token $x_0$. The coarse-scale encoder processes the instruction embedding $\hat{\ell}$ and viewpoint representations $X = [x_0, x_1, \ldots, x_{|\mathcal{V}_t|}]$ through cross-modal attention and Graph-Aware Self-Attention (GASA):

\small
\begin{equation}
    \text{GASA}(X) = \text{Softmax}\left(\frac{XW_q(XW_k)^T}{\sqrt{d}} + M\right)XW_v,
\label{eq:gasa}
\end{equation}
\normalsize

\noindent where the distance encoding matrix $M = EW_e + b_e$ incorporates the pairwise distance matrix $E$. For fine-scale planning, the fine-scale encoder processes the instruction $\hat{\ell}$ and panoramic features $r_t$ to generate action scores for immediate neighbors $\mathcal{N}(v_t)$. The final navigation decision combines both scales through learned dynamic weighting, producing action scores for each candidate viewpoint.

\subsection{Contrastive Variational World Model}

World models provide latent representations of environment dynamics, enabling efficient inference about future states. Given an observation sequence $(o_1, o_2, \ldots, o_T)$, the world model operates on latent states $z_t$ that capture environmental dynamics. The joint distribution factorizes as:

\begin{equation}
    \p(o, z) = \prod_{t=1}^{T} \p(z_t|z_{t-1}) \p(o_t|z_t).
\end{equation}

\noindent To maximize the observation likelihood $p(o_{1:T})$, the model introduces a variational posterior $q(z_{1:T}|o_{1:T})$ and derive the evidence lower bound (ELBO) \cite{2013vae}:

\small
\begin{equation}
\begin{aligned}
\ln \p(o) &\geq\sum_{t=1}^T \Big(\E{\q(z_t|o_{\leq t})}{\underbrace{\ln \p(o_t|z_t)}_{\mathcal{J}_{\mathrm{RECOVER}}}} \\
&- \E{q(z_{t-1}|o_{\leq t})}{\underbrace{\KL{\q(z_t|o_{\leq t})}{\p(z_t|z_{t-1})}}_{\mathcal{J}_{\mathrm{KL}}}} \Big).
\end{aligned}
\label{eq:basic_elbo}
\end{equation}
\normalsize

\noindent To empower the model with discriminative power while avoiding pixel-level reconstruction, the term $\mathcal{J}_{\mathrm{RECOVER}}$ is replaced with a contrastive objective \cite{2020dreamtocontrol,2021dreaming}. Following the information-theoretic derivation, we can lower-bound $\mathcal{J}_{\mathrm{RECOVER}}$ using noise-contrastive estimation (NCE) \cite{2018infonce}:

\small
\begin{equation}
\begin{aligned}
\mathcal{J}_{\mathrm{RECOVER}}
&\geq \E[\bigg]{\q(z_t|\cdot)}{\ln \p(z_t|o_t) -\ln \sum_{o' \in \mathcal{D}} \p(z_t|o')} \\
&= \mathcal{J}_{\mathrm{NCE}},
\end{aligned}
\label{eq:nce}
\end{equation}
\normalsize

\noindent where $\mathcal{D}$ represents a mini-batch of negative samples. This contrastive objective trains the model to distinguish between correct state-observation pairs $(z_t, o_t)$ and incorrect pairs $(z_t, o')$, effectively learning representations that capture environmental detail without explicit reconstruction. 

\section{Memoir}

This section presents Memoir, a memory-persistent VLN agent that employs world model imagination for adaptive experience retrieval. As illustrated in \Cref{fig:method}, our approach comprises three components: a language-conditioned contrastive world model that encodes histories and imagines future states as retrieval queries, a Hybrid Viewpoint-Level Memory (HVM) that stores both environmental observations and navigation histories for retrieval, and an experience-augmented navigation model integrating retrieved knowledge for navigation planning. To facilitate reading, we list the crucial notations in Memoir in \Cref{tab:notation}.

\begin{table}[t]
\centering
\caption{Key notation summary.}
\resizebox{0.48\textwidth}{!}{
{\renewcommand{\arraystretch}{1.15}
\begin{tabular}{ll}
\toprule
\textbf{Notation} & \textbf{Description} \\ 
\midrule
$\mathcal{G}_t = (\mathcal{V}_t, \mathcal{E}_t)$ & Episodic graph at step $t$ (viewpoints, edges) \\
$\mathcal{G}^{(k)} = (\mathcal{V}^{(k)}, \mathcal{E}^{(k)})$ & Persistent graph accumulated over $k$ episodes \\
$\ell$, $\hat{\ell}$ & Natural language instruction \& embedding \\
$o_t = \{o_t^{(i)}\}_{i=1}^{36}$, $r_t = \{r_t^{(i)}\}_{i=1}^{36}$ & Panoramic observation \& features (36 views) \\
$x_t=\text{AvgPool}(r_t)$& Viewpoint feature via average pooling \\
$\gamma_t$, $\epsilon$ & Reward signal (distance to goal), stop threshold \\
$z_t$, $\hat{z}_t$ & Inferred state \& imagined state \\
$\psi_s$, $\psi_o$ & State \& observation embedding functions \\
$\tau_t = \{\hat{z}_{t+i}\}_{i=1}^{H_t}$ & Imagined trajectory with horizon $H_t$ \\
$D$ & Overshooting dist. and max imagination horizon \\
$\mathcal{M}_o = (\mathcal{V}_o, \mathcal{X}_o)$ & Observation bank (viewpoints, features) \\
$\mathcal{M}_h = (\mathcal{V}_h, \mathcal{Z}_h, \mathcal{T}_h)$ & History bank (viewpoints, states, trajectories) \\
$c_{i,j}$, $c_i$ & Compatibility scores for retrieval \\
$W$, $P$ & Max width for obs. \& max patterns for history \\
$\rho_o$, $\gamma_o$ & Filter rate \& decay factor for obs. retrieval \\
$\theta_h$, $\gamma_h$ & Base threshold \& decay factor for history retrieval \\
$\sigma_c$, $\sigma_f$, $\sigma_h$ & Fusion weights for navigation model encoders \\
$s_j^{(c)}$, $s_j^{(f)}$, $s_j^{(h)}$ & Action scores (coarse, fine, history) \\
\bottomrule
\end{tabular}}}
\label{tab:notation}
\end{table}

\subsection{Language-Conditioned World Model}
To adapt the basic contrastive world model that focuses solely on environmental dynamics \cite{2020dreamtocontrol} for VLN task, our approach explicitly incorporates instruction conditioning to leverage the strong prior knowledge inherent in VLN tasks. We extend the standard ELBO formulation by incorporating instruction $\ell$ and reward signal $\gamma_t$ (indicating distance to goal):

\small
\begin{equation}
\begin{aligned}
\ln \p(o,\gamma|\ell)
&\geq\sum_{t=1}^T \Big(\E{\q(z_t|o_{\leq t},\ell)}{\underbrace{\ln \p(\gamma_t|z_t)}_{\mathcal{J}_{\mathrm{REWARD}}}} \\
+ & \E{\q(z_t|o_{\leq t},\ell)}{\underbrace{\ln \p(z_t|o_t) -\ln \sum_{o' \in \mathcal{D}} \p(z_t|o')}_{\mathcal{J}_{\mathrm{NCE}}}} \\
- & \E{q(z_{t-1}|o_{\leq t},\ell)}{\underbrace{\KL{\q(z_t|o_{\leq t},\ell)}{\p(z_t|z_{t-1})}}_{\mathcal{J}_{\mathrm{KL}}}} \Big),
\end{aligned}
\label{eq:elbo}
\end{equation}
\normalsize

\noindent where $\mathcal{J}_{\text{REWARD}}$ encourages accurate goal proximity prediction for imagination termination. The negative sample set $\mathcal{D}$ comprises observations from different timesteps and episodes within each training batch. The contrastive term $\mathcal{J}_{\text{NCE}}$ is implemented through a learnable function that measures compatibility between latent states and visual observations:

\begin{equation}
\begin{aligned}
    f(&z_t, o_t) = \frac{1}{\zeta} \operatorname{sim}(\psi_s(z_t), \psi_o(x_t)) \\
    & \p(z_t|o_t) \propto \exp(f(z_t, o_t)),
\end{aligned}
\label{eq:sim}
\end{equation}

\noindent where $x_t$ represents the visual feature extracted through DUET's observation encoder via averge pooling, $\psi_s$ and $\psi_o$ are learned embedding functions that map states and observations to a shared embedding space, $\text{sim}(a,b) = \frac{a^{\top}b}{\|a\|\|b\|}$ denotes cosine similarity, and $\zeta$ denotes temperature parameter. This formulation enables principled assessment of compatibility between imagined states and observations stored in long-term memory, providing a foundation for similarity-based memory retrieval.

\begin{algorithm}[t]
\SetEndCharOfAlgoLine{}
\SetKwComment{Comment}{// }{}
\SetKwInOut{Input}{Input}
\Input{
\begin{tabular}[t]{l @{\hspace{.1em}} l}%
$W$ & Max Width \\
$\rho_o$ & Filter Rate \\
$\gamma_o$ & Decay Factor \\
$\mathcal{G}^{(k)}$ & Persistent Graph \\
\end{tabular}%
\begin{tabular}[t]{l @{\hspace{.1em}} l}%
$v_t$ & Current Viewpoint \\
$\tau_t$ & Imagined States \\
$\mathcal{M}_o$ & Observation Bank \\
$\mathcal{G}_t$ & Episodic Graph \\
\end{tabular}%
}
\BlankLine
Initialize retrieval set $\mathcal{R} \gets \emptyset$ \\
\BlankLine
\For{$i \leftarrow 1$ \KwTo $|\tau_t|$}{
  Initialize $\mathcal{R}_{\text{tmp}} \gets \emptyset$ \\
  Get $i$-th order neighbors $\mathcal{N}_i(v_t)$ from $\mathcal{G}^{(k)}$ \\
  \For{each viewpoint $v_n \in \mathcal{N}_i(v_t)$}{
    Extract imagined state $\hat{z}_{t+i}$ from $\tau_t$ \\
    Compute compatibility $c_{i,n}$ via \Cref{eq:score_obs} \\
    Add $(v_n, c_{i,n})$ to $\mathcal{R}_{\text{tmp}}$ \\
  }
  Sort $\mathcal{R}_{\text{tmp}}$ by score $c_{i,n}$ in descending order \\
  Retain top $(1-\rho_o \cdot \gamma_o^{i-1})$ fraction of $\mathcal{R}_{\text{tmp}}$ \\
  Keep top $W$ nodes in $\mathcal{R}_{\text{tmp}}$ \\
  \For{each $(v_n, c_{i,n}) \in \mathcal{R}_{\text{tmp}}$}{
    Find shortest path $P_{t,n}$ from $v_t$ to $v_n$ in $\mathcal{G}^{(k)}$ \\
    Add path viewpoints: $\mathcal{R} \gets \mathcal{R} \cup P_{t,n}$ \\
  }
}
\BlankLine
\For{each viewpoint $v_n \in \mathcal{R}$}{
  Retrieve feature $x_n$ from $\mathcal{M}_o$ for viewpoint $v_n$ \\
  Retrieve edges $E_n$ from $\mathcal{G}^{(k)}$ for $v_n$ \\
  Update episodic graph: $\mathcal{G}_t.\text{update}(v_n, E_n)$ \\
  Store feature $x_n$ for viewpoint $v_n$ \\
}
\BlankLine
\Return updated episodic graph $\mathcal{G}_t$
\caption{Environmental Observation Retrieval}
\label{alg:obs}
\end{algorithm}

To improve the model's long-horizon predictive capability and enhance memory retrieval quality, we extend the ELBO formulation with multi-step overshooting. The $d$-step overshooting objective encourages accurate prediction over extended horizons:

\small
\begin{equation}
\begin{aligned}
&\quad\quad \mathcal{J}^{(d)} =\sum_{t=1}^T \Big(\E{\p(z_t|z_{t-d+1})\q(z_{t-d+1}|\cdot)}{\underbrace{\ln \p(\gamma_t|z_t)}_{\mathcal{J}_{\mathrm{REWARD}}}} + \\
&\E{\p(z_t|z_{t-d+1})\q(z_{t-d+1}|\cdot)}{\underbrace{\ln \p(z_t|o_t) -\ln \sum_{o'} \p(z_t|o')}_{\mathcal{J}_{\mathrm{NCE}}}} - \\
&\E{\p(z_{t-1}|z_{t-d}) \q(z_{t-d}|\cdot)}{\underbrace{\KL{\q(z_t|o_{\leq t},\ell)}{\p(z_t|z_{t-1})}}_{\mathcal{J}_{\mathrm{KL}}}}\Big).
\end{aligned}
\label{eq:overshoot}
\end{equation}
\normalsize

\noindent With maximum overshooting distance $D$, the final optimization objective becomes:

\small
\begin{equation}
\mathcal{J} = \mathcal{J}^{(1)} + \frac{1}{D-1} \sum_{d=2}^{D} \mathcal{J}^{(d)}.
\label{eq:objective}
\end{equation}
\normalsize

To efficiently optimize the objective in \Cref{eq:objective}, we adopt the Recurrent State-Space Model (RSSM) architecture \cite{2019rssm}, comprising four components:

\begin{equation}
    \begin{aligned}
        & \text{Inference Model:} && z_t \sim \q(z_t|z_{t-1}, o_t, \ell) \\
        & \text{Transition Model:} && \hat{z}_t \sim \p(z_t|z_{t-1}) \\
        & \text{Compatibility Model:} && \p(z_t|o_t) \propto \exp(f(z_t, o_t)) \\
        & \text{Reward Model:} && \hat{\gamma_t} \sim \p(\gamma_t|z_t), \\
    \end{aligned}
    \label{eq:world}
\end{equation}

\noindent where in practice the inference model takes $x_t$ as input for observation, and $\hat{\ell}$ as input for instruction. The inference model encodes navigation histories into representations for storage, and the transition model generates imagined future states that facilitate similarity-based memory retrieval.

\subsection{Hybrid Viewpoint-Level Memory (HVM)}

Having established how our world model imagines and infers states, we now describe how the imagined states query long-term memory. We introduce a dual-bank memory architecture that maintains both environmental observations and navigation behavioral histories at viewpoint granularity. HVM comprises two complementary banks organized around a persistent graph $\mathcal{G}^{(k)} = (\mathcal{V}^{(k)}, \mathcal{E}^{(k)})$ accumulated over $k$ episodes:

\begin{itemize}
\item \textbf{Observation Bank}: $\mathcal{M}_o = (\mathcal{V}_o, \mathcal{X}_o)$, where $\mathcal{V}_o = \{v_j\}$ represents the set of recorded viewpoints and $\mathcal{X}_o = \{x_j\}_{j=1}^{|\mathcal{V}_o|}$ contains corresponding viewpoint features extracted by DUET's observation encoder.

\item \textbf{History Bank}: $\mathcal{M}_h = (\mathcal{V}_h, \mathcal{Z}_h, \mathcal{T}_h)$, where $\mathcal{V}_h = \{v_j\}$ denotes viewpoints with recorded navigation histories, $\mathcal{Z}_h = \{\{z_j^{(k)}\}_{k=1}^{N_j}\}_{j=1}^{|\mathcal{V}_h|}$ stores inferred agent states from past episodes, and $\mathcal{T}_h = \{\{\tau_j^{(k)}\}_{k=1}^{N_j}\}_{j=1}^{|\mathcal{V}_h|}$ contains corresponding imagined trajectory sequences, where $N_j$ denotes the number of historical visits to viewpoint $v_j$.
\end{itemize}

At each timestep $t$, both memory banks are updated based on $v_t$: $\mathcal{M}_o$ receives the viewpoint feature $x_t$ extracted from observation $o_t$, while $\mathcal{M}_h$ stores the inferred state $z_t$ from the inference model and the imagined trajectory $\tau_t = \{\hat{z}_{t+i}\}_{i=1}^{H_t}$ generated by the transition model in \Cref{eq:world}. $H_t$ represents the imagination horizon, terminating when the predicted distance $\hat{\gamma}_{t+i}$ falls below threshold $\epsilon$ or reaches maximum horizon $D$.

\noindent\textbf{Environmental Observation Retrieval.} Given an imagined trajectory $\tau_t = \{\hat{z}_{t+i}\}_{i=1}^{H_t}$ at viewpoint $v_t$, we retrieve observations through topology-guided searching via state-observation compatibility. For imagined state $\hat{z}_{t+i}$ and stored feature $x_j$ at viewpoint $v_j$ from $\mathcal{M}_o$, we compute a compatibility score:

\begin{equation}
    c_{i,j} = \frac{1}{2} (\operatorname{sim}(\psi_s(\hat{z}_{t+i}), \psi_o(x_j))+1).
\label{eq:score_obs}
\end{equation}

This scoring mechanism directly leverages the contrastive objective from \Cref{eq:sim}, ensuring consistency between training and retrieval. For each imagination step $i$ and corresponding neighborhood order, the algorithm identifies all viewpoints in the $i$-th order neighborhood $\mathcal{N}_i(v_t) = \{v \in \mathcal{V}^{(k)} : d(v_t, v) = i\}$ and computes compatibility scores using \Cref{eq:score_obs}. Percentile-based filtering retains the top $(1-\rho_o \cdot \gamma_o^{i-1})$ fraction of viewpoints ranked by score, followed by selecting the top-$W$ viewpoints from the retained set. Finally, shortest paths from $v_t$ to all selected viewpoints are added to the episodic graph $\mathcal{G}_t$. The complete procedure is detailed in \Cref{alg:obs}.

\noindent\textbf{Navigation History Retrieval.} History retrieval identifies stored historical navigation patterns that exhibit similar imagined trajectories to the current agent's imagination. This process leverages the insight that agents with similar future expectations likely share comparable strategies and should benefit from each other's experiences. For a stored trajectory $\tau' = \{\hat{z}'_i\}_{i=1}^{H'}$ at viewpoint $v_t$ from the history bank $\mathcal{M}_h$, we perform sequential similarity matching based on imagined trajectory $\tau_t = \{\hat{z}_{t+i}\}_{i=1}^{H_t}$. The compatibility between imagined states at step $i$ is computed as:

\begin{equation}
    c_{i} = \frac{1}{2} (\operatorname{sim}(\psi_s(\hat{z}_{t+i}), \psi_s(\hat{z}'_i))+1).
\label{eq:score_nav}
\end{equation}

\begin{algorithm}[t]
\SetEndCharOfAlgoLine{}
\SetKwComment{Comment}{// }{}
\SetKwInOut{Input}{Input}
\Input{
\begin{tabular}[t]{l @{\hspace{.1em}} l}%
$P$ & Max Patterns \\
$\theta_h$ & Threshold \\
$\gamma_h$ & Decay Factor \\
$\mathcal{G}^{(k)}$ & Persistent Graph \\
\end{tabular}%
\begin{tabular}[t]{l @{\hspace{.1em}} l}%
$v_t$ & Current Viewpoint \\
$\tau_t$ & Imagined States \\
$\mathcal{M}_h$ & History Bank \\
$\mathcal{G}_t$ & Episodic Graph \\
\end{tabular}%
}
\BlankLine
Retrieve all patterns $Q$ from $\mathcal{M}_h$ for viewpoint $v_t$ \\
\BlankLine
\For{each $(z', \tau') \in Q$}{
  Initialize $L \gets \min(|\tau_t|, |\tau'|)$, scores $\mathcal{C} \gets \emptyset$ \\
  \For{$i \leftarrow 1$ \KwTo $L$}{
    Get imagined state $\hat{z}_{t+i}$, $\hat{z}'_i$ from $\tau_t$, $\tau'$ \\
    Compute compatibility $c_i$ via \Cref{eq:score_nav} \\
    \If{$c_i < \theta_h \cdot \gamma_h^{i-1}$}{
      \textbf{break}
    }
    Append score: $\mathcal{C} \gets \mathcal{C} \cup \{c_i\}$ \\
  }
  Store pattern with score: $(z', \tau', \mathcal{C}) \in Q$ \\
}
\BlankLine
Sort $Q$ in descending order (by length and score) \\
Retain top $P$ patterns as $Q$ \\
\BlankLine
\For{each $(z', \tau', \mathcal{C}) \in Q$}{
  Trace subsequent trajectory from $\mathcal{M}_h$: $\{z'_i, v_i\}_{i=1}^{|\mathcal{C}|}$ \\
  \For{$i \leftarrow 1$ \KwTo $|\mathcal{C}|$}{
    Retrieve feature $x_i$ from $\mathcal{M}_o$ for viewpoint $v_i$ \\
    Retrieve edges $E_i$ from $\mathcal{G}^{(k)}$ for $v_i$ \\
    Update episodic graph: $\mathcal{G}_t.\text{update}(v_i, E_i)$ \\
    Store state $z'_i$, score $c_i$ and $x_i$ for viewpoint $v_i$ \\
  }
}
\BlankLine
\Return updated episode graph $\mathcal{G}_t$
\caption{Navigation History Retrieval}
\label{alg:nav}
\end{algorithm}

For each imagination step $i$ in trajectory, we continue matching until either reaching the minimum of the two trajectory lengths, or encountering a compatibility score below the step-dependent threshold $\theta_h \cdot \gamma_h^{i-1}$. The compatibility scores up to the matching termination are stored as $\mathcal{C}$.

We rank stored trajectories using a two-stage criterion: matching length (longer matches preferred), and the minimum compatibility score among matched steps. The top-$P$ trajectory patterns are selected. For each selected pattern, we retrieve the subsequent $|\mathcal{C}|$ viewpoints and their associated inferred states $\{z'_i, v_i\}_{i=1}^{|\mathcal{C}|}$, incorporating the corresponding subgraph structure into $\mathcal{G}_t$ along with state representations and compatibility scores. The complete procedure is outlined in \Cref{alg:nav}.

\subsection{Navigation Model}

At each timestep $t$, the agent imagines future states $\tau_t$ and retrieve environmental observation and navigation history according to $v_t$. The retrieved information is then integrated into the episodic topological graph $\mathcal{G}_t$ maintained by topological mapping. Now, we extend DUET \cite{2022duet} with specialized processing encoders that integrate retrieved experiential knowledge into navigation decisions. Our model processes these retrieved information through dedicated encoders: global observations, local observations and navigation behavioral patterns. The navigation model comprises three branches:

\noindent\textbf{Coarse-Scale Encoder.} The coarse-scale encoder incorporates retrieved observations by expanding viewpoint representations $X$ with an additional type—retrieved viewpoints. The full viewpoint representations $X = [x_0, x_1, \ldots, x_{|\mathcal{V}_t|}]$ containing retrieved observations are processed through the coarse-scale encoder for $\hat{X}$. Global action scores are computed as $s_j^{(c)} = \text{FFN}(\hat{x}_j)$ for viewpoint $v_j$, providing high-level navigation preferences.

\noindent\textbf{Fine-Scale Encoder.} The fine-scale encoder processes immediate panoramic feature $r_t$ for $\hat{r}_t$. Local action scores $s_j^{(f)} = \text{FFN}(\hat{r}_t^{(i_j)})$ are computed for each neighbor $v_j \in \mathcal{N}(v_t)$ and converted to the global action space:

\begin{equation}
    s_j^{(f')} = \begin{cases}
    s_{\text{back}}, & \text{if } v_j \in \mathcal{V}_t \setminus \mathcal{N}(v_t) \\
    s_j^{(f)}, & \text{otherwise},
    \end{cases}
\end{equation}

\noindent where $i_j$ denotes the view index oriented toward $v_j$ and $s_{\text{back}}$ aggregates scores for all visited neighbors of viewpoint $v_t$ to encourage backtracking when necessary.

\noindent\textbf{Navigation-History Encoder.} The navigation-history encoder processes retrieved behavioral patterns by fusing historical states with current viewpoint representations. The node set $\mathcal{V}_h$ includes all viewpoints processed by this branch, comprising both currently visited locations and nodes retrieved from the history bank. For each viewpoint $v_j \in \mathcal{V}_h$ with retrieved states $Z_j = [z'^{(1)}_j, z'^{(2)}_j, \ldots, z'^{(N_j)}_j]$ and compatibility scores $C_j = [c^{(1)}_j, c^{(2)}_j, \ldots, c^{(N_j)}_j]$, where $N_j$ denotes the number of retrieved states at $v_j$, we compute:

\small
\begin{equation}
    u_j = \left(\operatorname{softmax}\left(\frac{C_j}{\zeta}\right)\right)^{\top} Z_j + x_j.
\label{eq:softmax}
\end{equation}
\normalsize

For visited nodes without retrieved historical states, we simply use the observation $u_j = x_j$. The fused state representations $U = [u_1, u_2, \ldots, u_{|\mathcal{V}_h|}]$ are processed through a transformer to produce history-informed action scores $s_i^{(h)}$, which are then mapped to the global action space:

\begin{equation}
    s_i^{(h')} = \begin{cases}
    s_0, & \text{if } v_i \in \mathcal{V}_t \setminus \mathcal{V}_h \\
    s_i^{(h)}, & \text{otherwise}.
    \end{cases}
\end{equation}

\noindent\textbf{Dynamic Fusion.} We implement a learned dynamic fusion mechanism that automatically balances contributions from the three branches based on current situational factors. The fusion weights are computed through:

\begin{equation}
    [\sigma_f, \sigma_c, \sigma_h] = \text{Softmax}(\text{FFN}([\hat{r}_0; \hat{x}_0; \hat{u}_0])),
\end{equation}

\noindent where $\hat{r}_0$, $\hat{x}_0$, and $\hat{u}_0$ represent the encoded stop token representations from fine-scale, coarse-scale, and navigation-history encoders respectively, and $[;]$ denotes concatenation. The final navigation scores integrate all three branches:

\begin{equation}
\begin{aligned}
    s_j = \sigma_f s_j^{(f')} + \sigma_c s_j^{(c)} + \sigma_h s_j^{(h')}.
\end{aligned}
\end{equation}

As presented in \Cref{alg:method}, Memoir realizes imagination-guided memory retrieval by generating imagined trajectories as queries to adaptively access relevant observations and behavioral histories from persistent memory. The navigation model integrates retrieved experiences, enabling informed decisions grounded by historical evidence while continuously updating memory banks for progressive improvement across episodes.

\begin{algorithm}[t]
\SetEndCharOfAlgoLine{}
\SetKwComment{Comment}{// }{}
\SetKwInOut{Input}{Input}
\Input{\\\hspace{-3.6em}\small
\begin{tabular}[t]{l @{\hspace{.1em}} l}%
$T_{\max}$ & Max Step limit \\
$D$ & Imagination Horizon \\
$\epsilon$ & Stop threshold \\
\end{tabular}\hspace{-0.5em}%
\begin{tabular}[t]{l @{\hspace{.1em}} l}%
$\mathcal{M}_o$ & Observation Bank \\
$\mathcal{M}_h$ & History Bank \\
$\mathcal{G}^{(k)}$ & Persistent Graph \\
\end{tabular}%
}
\BlankLine
Initialize episodic graph $\mathcal{G}_0 \leftarrow \emptyset$ \\
Receive initial observation $o_1$, viewpoint $v_1$ \\
\BlankLine
\For{step $t = 1$ \KwTo $T_{\max}$}{
  Update topological graphs $\mathcal{G}_t$ and $\mathcal{G}^{(k)}$ \\
  Infer current state $z_t \sim \q(z_t|z_{t-1},o_t,\ell)$ \\
  Initialize imagined trajectory $\tau_t \leftarrow \emptyset$ \\
  \For{$i = 1$ \KwTo $D$}{
    Imagine next state $\hat{z}_{t+i} \sim \p(z_{t+i}|z_{t+i-1})$ \\
    Predict reward $\hat{\gamma}_{t+i} \sim \p(\gamma_{t+i}|z_{t+i})$ \\
    Update trajectory $\tau_t \leftarrow \tau_t \cup \{\hat{z}_{t+i}\}$ \\
    \If{$\hat{\gamma}_{t+i} < \epsilon$}{
      break
    }
  }
  \BlankLine
  $\mathcal{G}_t \leftarrow$ ObsRetrieval($\mathcal{M}_o, v_t, \tau_t, \mathcal{G}_{t}, \mathcal{G}^{(k)}$) \Comment{Algorithm \ref{alg:obs}}
  $\mathcal{G}_t \leftarrow$ HistoryRetrieval($\mathcal{M}_h, v_t, \tau_t, \mathcal{G}_t, \mathcal{G}^{(k)}$) \Comment{Algorithm \ref{alg:nav}}
  \BlankLine
  Extract viewpoint feature $x_t$ from $o_t$ \\
  $\mathcal{M}_o.\text{add}(v_t, x_t)$ \\
  $\mathcal{M}_h.\text{add}(v_t, z_t, \tau_t)$ \\
  \BlankLine
  Compute score $s_j$ for each candidate node $v_j$ \\
  Select action $a_t \leftarrow \argmax_j s_j$ \\
  
  \If{$a_t = \text{stop}$}{
    break
  }
  Receive $o_{t+1}$, $v_{t+1} \leftarrow \text{env.step}(a_t)$ \\
}
\caption{Memoir Navigation Loop}
\label{alg:method}
\end{algorithm}

\setlength{\tabcolsep}{.3em}
\section{Experiments}
\subsection{Experimental Setup}

\textbf{Datasets.}
We evaluate Memoir on two established memory-persistent VLN benchmarks that provide complementary evaluation perspectives. Iterative Room-to-Room (IR2R) \cite{2023ivln} extends the foundational Room-to-Room (R2R) dataset \cite{2018r2r} to multi-episode scenarios through structured tours, containing 183 training tours with an average length of 76.6 episodes. The validation splits comprise seen environments (159 tours, average 6.4 episodes) and unseen environments (33 tours, average 71.2 episodes). General Scene Adaptation (GSA-R2R) \cite{2025gsa} incorporates 150 Habitat-Matterport3D (HM3D) scenes \cite{2021hm3d} with 600 paths per scene, providing 90,000 total episodes across 10 evaluation scenarios covering residential and non-residential environments with various instruction types including basic navigational instructions, scene-specific instructions, and user-personalized instructions.

\noindent\textbf{Implementation Details.}
We implement Memoir on three foundational models: DUET \cite{2022duet} and ScaleVLN \cite{2023scalevln} representing traditional VLN models, and GR-DUET \cite{2025gsa} representing the memory-persistent approaches. All models utilize pretrained weights from their respective pretraining phases without task-specific fine-tuning. For rigorous comparison, we retrain all baseline models with identical hyperparameters and experimental conditions, including synchronized episode ordering in GSA-R2R. Our world model implementation employs two architectural variants: GRU and Transformer. Both variants utilize textual embeddings and share the observation encoder with the navigation model. Joint pretraining of the world model and navigation model is conducted on R2R and augmented trajectories \cite{2020prevalent} for 5,000 iterations with batch size 32 and learning rate 5e-5, followed by imitation learning at learning rate 1e-5. All results are reported over 3 separate runs.

\noindent\textbf{Evaluation Metrics.}
We employ standard VLN metrics \cite{2018evaluation} for navigation performance evaluation. To quantify the effectiveness of long-term memory retrieval, we introduce four complementary metrics that evaluate both observation retrieval and history retrieval quality. The metrics include:

\textit{Trajectory Length (TL)}: predicted path length in meters.

\textit{Navigation Error (NE)}: distance between agent's final position to target in meters.

\textit{Success Rate (SR)}: the percentage of final positions less than 3 meters away from the target location.

\textit{Success Rate penalized by Path Length (SPL)}: SR normalized by the ratio between the length of the shortest path and the predicted path.

\textit{Normalized Dynamic Time Warping (nDTW)}: dynamic time warping normalized between predicted and expert paths.

\textit{Tour-normalized Dynamic Time Warping (T-nDTW)}: the overall navigation consistency across complete tours.

\textit{Observation Accuracy (OA)}: the precision of retrieved observations from the observation bank $\mathcal{M}_o$ across the episode:
\begin{equation}
\text{OA} = \frac{|\bigcup_{t=1}^{T}(\mathcal{R}_t \cap \mathcal{V}_{\text{gt},t}^o)|}{|\bigcup_{t=1}^{T}\mathcal{R}_t|},
\end{equation}
where $\mathcal{R}_t$ denotes viewpoints retrieved from $\mathcal{M}_o$ at timestep $t$ (as detailed in Algorithm~\ref{alg:obs}), and $\mathcal{V}_{\text{gt},t}^o$ represents the ground truth viewpoints on the teacher trajectory within $D$ steps that exist in the observation bank, where $\mathcal{V}_o$ denotes all viewpoints stored in $\mathcal{M}_o$ and $T$ is episode length.

\textit{Observation Recall (OR)}: the coverage of relevant environmental observations across the episode:
\begin{equation}
\text{OR} = \frac{|\bigcup_{t=1}^{T}(\mathcal{R}_t \cap \mathcal{V}_{\text{gt},t}^o)|}{|\bigcup_{t=1}^{T}\mathcal{V}_{\text{gt},t}^o|}.
\end{equation}

\textit{History Accuracy (HA)}: the precision of retrieved navigation patterns from the history bank $\mathcal{M}_h$ across the episode:
\begin{equation}
\text{HA} = \frac{\sum_{t=1}^{T}\sum_{j=1}^{|Q_t|}|\mathcal{V}_{\text{traj},t,j}^h \cap \mathcal{V}_{\text{gt},t,j}^h|}{\sum_{t=1}^{T}\sum_{j=1}^{|Q_t|}|\mathcal{V}_{\text{traj},t,j}^h|},
\end{equation}
where $|Q_t|$ denotes the number of retrieved navigation history patterns at timestep $t$ (as detailed in Algorithm~\ref{alg:nav}), and $\mathcal{V}_{\text{traj},t,j}^h = \{v_1^{(j)}, v_2^{(j)}, \ldots, v_{|\mathcal{C}_j|}^{(j)}\}$ represents the sequence of viewpoints in the $j$-th retrieved navigation history trajectory, and $\mathcal{V}_{\text{gt},t,j}^h$ represents the viewpoints on the teacher trajectory that exist in the original history trajectory.

\textit{History Recall (HR)}: the coverage of relevant navigation patterns across the episode:
\begin{equation}
\text{HR} = \frac{\sum_{t=1}^{T}\sum_{j=1}^{|Q_t|}|\mathcal{V}_{\text{traj},t,j}^h \cap \mathcal{V}_{\text{gt},t,j}^h|}{\sum_{t=1}^{T}\sum_{j=1}^{|Q_t|}|\mathcal{V}_{\text{gt},t,j}^h|}.
\end{equation}

\begin{table*}[t]
    \renewcommand{\arraystretch}{1.12}
    \setlength{\aboverulesep}{0pt}
    \setlength{\belowrulesep}{0pt}
    \setlength{\tabcolsep}{1.8pt}
    \centering
    \caption{
	    Comparison of navigation performance between Memoir and various VLN methods on the IR2R benchmark.
	}
	\resizebox{\textwidth}{!}{
    \begin{threeparttable}
		\begin{tabular}{lcccc c cccccc c cccccc}
			\toprule
            & & & & & &
            \multicolumn{6}{c}{\textbf{Val Seen}}
		   && \multicolumn{6}{c}{\textbf{Val Unseen}}
            \\
			\cmidrule{7-12}
			\cmidrule{14-19}
			Methods
			& \textsc{ph}
			& \textsc{th}
			& \textsc{phi}
			& \textsc{iw}
			&
            & \textbf{\texttt{TL}}~$\downarrow$
			& \textbf{\texttt{NE}}~$\downarrow$
			& \textbf{\texttt{nDTW}}~$\uparrow$
			& \textbf{\texttt{SR}}~$\uparrow$
			& \textbf{\texttt{SPL}}~$\uparrow$
			& \textbf{\texttt{t-nDTW}}~$\uparrow$
			&
			& \textbf{\texttt{TL}}~$\downarrow$
			& \textbf{\texttt{NE}}~$\downarrow$
			& \textbf{\texttt{nDTW}}~$\uparrow$
			& \textbf{\texttt{SR}}~$\uparrow$
			& \textbf{\texttt{SPL}}~$\uparrow$
			& \textbf{\texttt{t-nDTW}}~$\uparrow$
			\\
			\midrule
			\midrule
			HAMT \cite{2021hamt}
                & & & & &
                & 10.1 \scriptsize{$\pm$0.1}
                & 4.2 \scriptsize{$\pm$0.1}
                & 71 \scriptsize{$\pm$1}
                & 63 \scriptsize{$\pm$1}
                & 61 \scriptsize{$\pm$1}
                & 58 \scriptsize{$\pm$1}
                &
                & \bf \hphantom{0}9.4 \scriptsize{$\pm$0.1}
                & 4.7 \scriptsize{$\pm$0.0}
                & 66 \scriptsize{$\pm$0}
                & 56 \scriptsize{$\pm$0}
                & 54 \scriptsize{$\pm$0}
                & 50 \scriptsize{$\pm$0}
			\\
			TourHAMT \cite{2023ivln}
			    & \checkmark & \checkmark & \checkmark & \checkmark &
                & \bf \hphantom{0}9.4 \scriptsize{$\pm$0.4}
                & 5.8 \scriptsize{$\pm$0.1}
                & 59 \scriptsize{$\pm$0}
                & 45 \scriptsize{$\pm$1}
                & 43 \scriptsize{$\pm$1}
                & 45 \scriptsize{$\pm$0}
                &
                & 10.0 \scriptsize{$\pm$0.2}
                & 6.2 \scriptsize{$\pm$0.1}
                & 52 \scriptsize{$\pm$0}
                & 39 \scriptsize{$\pm$1}
                & 36 \scriptsize{$\pm$0}
                & 32 \scriptsize{$\pm$1}
			\\
			\scriptsize 
			    & \checkmark & \checkmark & \checkmark &  &
                & 10.5 \scriptsize{$\pm$0.3}
                & 6.0 \scriptsize{$\pm$0.2}
                & 58 \scriptsize{$\pm$1}
                & 45 \scriptsize{$\pm$2}
                & 43 \scriptsize{$\pm$2}
                & 42 \scriptsize{$\pm$1}
                &
                & 10.9 \scriptsize{$\pm$0.2}
                & 6.8 \scriptsize{$\pm$0.2}
                & 51 \scriptsize{$\pm$1}
                & 38 \scriptsize{$\pm$1}
                & 34 \scriptsize{$\pm$1}
                & 31 \scriptsize{$\pm$1}
			\\
			\scriptsize 
			    & \checkmark & \checkmark &  &  &
                & 10.6 \scriptsize{$\pm$0.3}
                & 6.0 \scriptsize{$\pm$0.1}
                & 58 \scriptsize{$\pm$1}
                & 45 \scriptsize{$\pm$1}
                & 42 \scriptsize{$\pm$1}
                & 42 \scriptsize{$\pm$1}
                &
                & 10.3 \scriptsize{$\pm$0.3}
                & 6.7 \scriptsize{$\pm$0.2}
                & 50 \scriptsize{$\pm$1}
                & 38 \scriptsize{$\pm$1}
                & 34 \scriptsize{$\pm$1}
                & 29 \scriptsize{$\pm$1}
			\\
			\scriptsize 
			    & \checkmark &  &  &  &
                & 10.9 \scriptsize{$\pm$0.3}
                & 6.1 \scriptsize{$\pm$0.1}
                & 58 \scriptsize{$\pm$1}
                & 45 \scriptsize{$\pm$1}
                & 42 \scriptsize{$\pm$1}
                & 41 \scriptsize{$\pm$0}
                &
                & 11.0 \scriptsize{$\pm$0.6}
                & 6.7 \scriptsize{$\pm$0.1}
                & 51 \scriptsize{$\pm$0}
                & 38 \scriptsize{$\pm$0}
                & 34 \scriptsize{$\pm$0}
                & 28 \scriptsize{$\pm$1}
			\\
                OVER-NAV \cite{2024overnav}
                & & & & &
                & 9.9 \scriptsize{$\pm$0.1}
                & 3.7 \scriptsize{$\pm$0.1}
                & 73 \scriptsize{$\pm$1}
                & 65 \scriptsize{$\pm$1}
                & 63 \scriptsize{$\pm$1}
                & 62 \scriptsize{$\pm$0}
                &
                & \bf \hphantom{0}9.4 \scriptsize{$\pm$0.1}
                & 4.1 \scriptsize{$\pm$0.1}
                & 69 \scriptsize{$\pm$0}
                & 60 \scriptsize{$\pm$1}
                & 57 \scriptsize{$\pm$0}
                & \bf 55 \scriptsize{$\pm$1}
            \\
            \midrule
            \multicolumn{19}{c}{\textbf{Comparison with Traditional VLN Models:}}\\
                \rowcolor{gray!15}\multicolumn{19}{l}{\emph{VLN models pretrained with default protocol:}}\\
                DUET \cite{2022duet}
                & & & & &

                & 12.5 \scriptsize{$\pm$0.4}
                & 2.2 \scriptsize{$\pm$0.1}
                & 79.8 \scriptsize{$\pm$1.1}
                & 79.8 \scriptsize{$\pm$0.7}
                & 74.5 \scriptsize{$\pm$0.9}
                & 69.1 \scriptsize{$\pm$1.7}
                &
                & 14.4 \scriptsize{$\pm$0.1}
                & 3.5 \scriptsize{$\pm$0.0}
                & 65.0 \scriptsize{$\pm$0.1}
                & 69.2 \scriptsize{$\pm$0.3}
                & 58.0 \scriptsize{$\pm$0.1}
                & 47.0 \scriptsize{$\pm$0.8}
            \\
                +Memoir (w/o retrieval)
                & & & & &
                & 11.2 \scriptsize{$\pm$0.0}
                & 2.3 \scriptsize{$\pm$0.0}
                & 81.2 \scriptsize{$\pm$0.0}
                & 79.4 \scriptsize{$\pm$0.4}
                & 75.8 \scriptsize{$\pm$0.8}
                & 72.3 \scriptsize{$\pm$0.3}
                &
                & 12.1 \scriptsize{$\pm$0.2}
                & 3.4 \scriptsize{$\pm$0.0}
                & 69.3 \scriptsize{$\pm$0.8}
                & 70.8 \scriptsize{$\pm$0.5}
                & 62.2 \scriptsize{$\pm$0.9}
                & 52.1 \scriptsize{$\pm$1.2}
            \\
                +Memoir (Ours)
                & & & & &
                & 11.5 \scriptsize{$\pm$0.1}
                & 2.6 \scriptsize{$\pm$0.2}
                & 78.9 \scriptsize{$\pm$0.9}
                & 77.1 \scriptsize{$\pm$0.5}
                & 72.8 \scriptsize{$\pm$0.5}
                & 68.0 \scriptsize{$\pm$0.8}
                &
                & \bf 11.0 \scriptsize{$\pm$0.0}
                & \bf 2.8 \scriptsize{$\pm$0.1}
                & \bf 75.2 \scriptsize{$\pm$0.0}
                & \bf 75.4 \scriptsize{$\pm$0.2}
                & \bf 69.1 \scriptsize{$\pm$0.3}
                & \bf 58.8 \scriptsize{$\pm$0.4}
			\\
                \rowcolor{gray!15}\multicolumn{19}{l}{\emph{VLN models pretrained with environmental augmentation:}}\\
                ScaleVLN \cite{2023scalevln}
                & & & & &
                & 12.8 \scriptsize{$\pm$0.0}
                & 2.2 \scriptsize{$\pm$0.0}
                & 79.6 \scriptsize{$\pm$0.4}
                & 79.5 \scriptsize{$\pm$0.5}
                & 74.1 \scriptsize{$\pm$0.6}
                & 67.0 \scriptsize{$\pm$0.2}
                &
                & 13.5 \scriptsize{$\pm$0.0}
                & 2.7 \scriptsize{$\pm$0.0}
                & 71.6 \scriptsize{$\pm$0.1}
                & 76.2 \scriptsize{$\pm$0.1}
                & 66.5 \scriptsize{$\pm$0.2}
                & 53.4 \scriptsize{$\pm$0.2}
			\\
                +Memoir (w/o retrieval)
                & & & & &
                & 12.4 \scriptsize{$\pm$0.1}
                & 2.4 \scriptsize{$\pm$0.1}
                & 79.2 \scriptsize{$\pm$0.4}
                & 79.1 \scriptsize{$\pm$0.5}
                & 74.1 \scriptsize{$\pm$0.2}
                & 67.4 \scriptsize{$\pm$1.8}
                &
                & 12.6 \scriptsize{$\pm$0.2}
                & 2.6 \scriptsize{$\pm$0.1}
                & 74.5 \scriptsize{$\pm$0.7}
                & 76.8 \scriptsize{$\pm$0.3}
                & 69.1 \scriptsize{$\pm$0.3}
                & 56.3 \scriptsize{$\pm$2.7}
			\\
                +Memoir (Ours)
                & & & & &
                & 11.6 \scriptsize{$\pm$0.2}
                & 2.5 \scriptsize{$\pm$0.1}
                & 78.7 \scriptsize{$\pm$0.1}
                & 76.1 \scriptsize{$\pm$0.5}
                & 72.3 \scriptsize{$\pm$0.1}
                & 67.1 \scriptsize{$\pm$0.0}
                &
                & \bf 10.9 \scriptsize{$\pm$0.2}
                & \bf 2.6 \scriptsize{$\pm$0.0}
                & \bf 77.2 \scriptsize{$\pm$0.6}
                & \bf 77.4 \scriptsize{$\pm$0.2}
                & \bf 72.1 \scriptsize{$\pm$0.4}
                & \bf 62.2 \scriptsize{$\pm$0.6}
			\\
            \midrule
            \multicolumn{19}{c}{\textbf{Comparison with Memory-Persistent VLN Models:}}\\
                \rowcolor{gray!15}\multicolumn{19}{l}{\emph{VLN models pretrained with full navigation graph:}}\\
                GR-DUET \cite{2025gsa}
                & & & & &
                & 12.5 \scriptsize{$\pm$0.7}
                & 4.3 \scriptsize{$\pm$0.2}
                & 65.2 \scriptsize{$\pm$0.7}
                & 61.1 \scriptsize{$\pm$1.4}
                & 55.1 \scriptsize{$\pm$0.1}
                & 49.0 \scriptsize{$\pm$0.5}
                &
                & 11.0 \scriptsize{$\pm$0.2}
                & 3.1 \scriptsize{$\pm$0.0}
                & 74.5 \scriptsize{$\pm$0.5}
                & 72.7 \scriptsize{$\pm$0.5}
                & 67.9 \scriptsize{$\pm$0.1}
                & 54.8 \scriptsize{$\pm$0.1}
            \\
                +Memoir (w/o retrieval)
                & & & & &
                & 12.9 \scriptsize{$\pm$0.1}
                & \bf 2.8 \scriptsize{$\pm$0.1}
                & \bf 75.6 \scriptsize{$\pm$0.0}
                & \bf 76.7 \scriptsize{$\pm$0.8}
                & \bf 70.1 \scriptsize{$\pm$0.3}
                & \bf 63.9 \scriptsize{$\pm$0.7}
                &
                & 12.5 \scriptsize{$\pm$0.3}
                & 3.2\scriptsize{$\pm$0.1}
                & 70.3 \scriptsize{$\pm$0.1}
                & 72.7 \scriptsize{$\pm$0.5}
                & 64.0 \scriptsize{$\pm$0.0}
                & 52.0 \scriptsize{$\pm$0.5}
			\\
                +Memoir (Ours)
                & & & & &
                & \bf 11.8 \scriptsize{$\pm$0.4}
                & 3.0 \scriptsize{$\pm$0.0}
                & 74.1 \scriptsize{$\pm$0.3}
                & 72.2 \scriptsize{$\pm$0.2}
                & 66.7 \scriptsize{$\pm$0.5}
                & 61.9 \scriptsize{$\pm$0.3}
                &
                & \bf 10.2 \scriptsize{$\pm$0.2}
                & \bf 2.5 \scriptsize{$\pm$0.0}
                & \bf 79.2 \scriptsize{$\pm$0.2}
                & \bf 77.6 \scriptsize{$\pm$0.5}
                & \bf 73.3 \scriptsize{$\pm$0.1}
                & \bf 66.9 \scriptsize{$\pm$0.7}
			\\
			\bottomrule
		\end{tabular}
        \begin{tablenotes}
	    	\footnotesize
            \item (w/o Retrieval): Variant pretrained and finetuned under identical conditions to Memoir, excluding only the explicit memory retrieval mechanism.
	    	\item \textsc{ph}: previous history integration; \textsc{th}: trainable history encoder; \textsc{phi}: previous history identifier; \textsc{iw}: inflection weighting.
	    \end{tablenotes}
    \end{threeparttable}}
	\label{tab:ir2r_results}
\end{table*}

\begin{table*}[t]
    \setlength{\tabcolsep}{3pt}
    \renewcommand{\arraystretch}{1.12}
    \setlength{\aboverulesep}{0pt}
    \setlength{\belowrulesep}{0pt}
    \setlength{\tabcolsep}{4pt}
    \centering
    \caption{
        Comprehensive comparison of navigation performance on the GSA-R2R benchmark.
    }
    \resizebox{\textwidth}{!}{
    \begin{threeparttable}
        \begin{tabular}{l c cc c ccc c ccc c ccc}
            \toprule
            & &
            \multicolumn{2}{c}{\textbf{User Instructions}}
            && \multicolumn{3}{c}{\textbf{Scene Instructions}}
            && \multicolumn{7}{c}{\textbf{Basic Instructions}}\\
            \cmidrule{3-4} \cmidrule{6-8} \cmidrule{10-16}
            & &
            \multicolumn{2}{c}{\textbf{Residential}}
            && \multicolumn{3}{c}{\textbf{Non-Residential}} && \multicolumn{3}{c}{\textbf{Residential}} && \multicolumn{3}{c}{\textbf{Non-Residential}}
            \\
            \cmidrule{3-4} \cmidrule{6-8} \cmidrule{10-12} \cmidrule{14-16}
            Methods
            &
            & \textbf{\texttt{SR}}~$\uparrow$
            & \textbf{\texttt{SPL}}~$\uparrow$
            &
            & \textbf{\texttt{SR}}~$\uparrow$
            & \textbf{\texttt{SPL}}~$\uparrow$
            & \textbf{\texttt{nDTW}}~$\uparrow$
            &
            & \textbf{\texttt{SR}}~$\uparrow$
            & \textbf{\texttt{SPL}}~$\uparrow$
            & \textbf{\texttt{nDTW}}~$\uparrow$
            &
            & \textbf{\texttt{SR}}~$\uparrow$
            & \textbf{\texttt{SPL}}~$\uparrow$
            & \textbf{\texttt{nDTW}}~$\uparrow$
            \\
            \midrule
            \midrule
                TourHAMT \cite{2023ivln}
                & & 14.7 & 12.0 & & 9.7 \scriptsize{$\pm$0.1} & 8.0 \scriptsize{$\pm$0.1} & 32.3 \scriptsize{$\pm$0.1} & & 14.9 \scriptsize{$\pm$0.1} & 12.2 \scriptsize{$\pm$0.1} & 34.7 \scriptsize{$\pm$0.1} & & 11.0 \scriptsize{$\pm$0.2} & 8.6 \scriptsize{$\pm$0.2} & 32.2 \scriptsize{$\pm$0.1} \\
                OVER-NAV \cite{2024overnav}
                & & 20.4 & 16.1 & & 16.7 \scriptsize{$\pm$0.4} & 12.6 \scriptsize{$\pm$0.2} & 34.6 \scriptsize{$\pm$0.3} & & 22.3 \scriptsize{$\pm$0.3} & 16.8 \scriptsize{$\pm$0.2} & 37.1 \scriptsize{$\pm$0.1} & & 16.6 \scriptsize{$\pm$0.2} & 13.0 \scriptsize{$\pm$0.1} & 35.0 \scriptsize{$\pm$0.2} \\
                \midrule
                DUET \cite{2022duet}
                & & 54.6 & 44.9 & & 39.6 & 30.1 & 40.9 & & 57.7 & 47.0 & 55.6 & & 48.1 & 37.3 & 45.9 \\
            \quad +MLM \cite{2020prevalent}
                & & 55.2 & 45.2 & & 39.8 \scriptsize{$\pm$0.1} & 30.5 \scriptsize{$\pm$0.1} & 41.1 \scriptsize{$\pm$0.1} & & 57.9 \scriptsize{$\pm$0.2} & 47.3 \scriptsize{$\pm$0.1} & 55.9 \scriptsize{$\pm$0.2} & & 48.3 \scriptsize{$\pm$0.5} & 38.8 \scriptsize{$\pm$0.5} & 48.4 \scriptsize{$\pm$0.3} \\
            \quad +MRC \cite{2020prevalent}
                & & 54.5 & 44.8 & & 39.7 \scriptsize{$\pm$0.1} & 30.2 \scriptsize{$\pm$0.1} & 40.9 \scriptsize{$\pm$0.1} & & 57.7 \scriptsize{$\pm$0.1} & 47.0 \scriptsize{$\pm$0.1} & 55.6 \scriptsize{$\pm$0.1} & & 48.1 \scriptsize{$\pm$0.1} & 37.3 \scriptsize{$\pm$0.1} & 45.9 \scriptsize{$\pm$0.1} \\
            \quad +BT \cite{2020bt}
                & & 59.0 & 55.7 & & 41.2 \scriptsize{$\pm$1.5} & 38.2 \scriptsize{$\pm$1.2} & 51.3 \scriptsize{$\pm$1.2} & & 61.3 \scriptsize{$\pm$0.6} & 57.7 \scriptsize{$\pm$0.3} & 70.1 \scriptsize{$\pm$0.5} & & 49.5 \scriptsize{$\pm$0.8} & 46.0 \scriptsize{$\pm$0.8} & 59.4 \scriptsize{$\pm$0.9} \\
            \quad +TENT \cite{2021tent}
                & & 53.8 & 42.3 & & 40.6 \scriptsize{$\pm$0.2} & 28.9 \scriptsize{$\pm$0.2} & 38.9 \scriptsize{$\pm$0.2} & & 57.2 \scriptsize{$\pm$0.4} & 44.2 \scriptsize{$\pm$0.4} & 52.9 \scriptsize{$\pm$0.1} & & 46.5 \scriptsize{$\pm$0.4} & 33.7 \scriptsize{$\pm$0.2} & 42.6 \scriptsize{$\pm$0.3} \\
            \quad +SAR \cite{2023tan}
                & & 53.7 & 41.9 & & 41.4 \scriptsize{$\pm$0.6} & 29.1 \scriptsize{$\pm$0.3} & 39.0 \scriptsize{$\pm$0.3} & & 57.6 \scriptsize{$\pm$0.2} & 44.6 \scriptsize{$\pm$0.2} & 53.0 \scriptsize{$\pm$0.2} & & 44.6 \scriptsize{$\pm$1.5} & 31.5 \scriptsize{$\pm$1.6} & 40.6 \scriptsize{$\pm$1.3} \\
                \midrule
                \rowcolor{gray!15}\multicolumn{16}{l}{\emph{VLN models pretrained with full navigation graph:}}\\
                GR-DUET \cite{2025gsa}
                & & 64.8 & 59.6 & & 48.1 \scriptsize{$\pm$0.1} & 42.8 \scriptsize{$\pm$0.1} & 53.7 \scriptsize{$\pm$0.1} & & 69.3 \scriptsize{$\pm$0.2} & 64.3 \scriptsize{$\pm$0.1} & 71.4 \scriptsize{$\pm$0.1} & & 56.6 \scriptsize{$\pm$0.1} & 51.5 \scriptsize{$\pm$0.1} & 61.0 \scriptsize{$\pm$0.1} \\
                GR-DUET* \cite{2025gsa}
                & & 63.8 & 59.8 & & 47.1 \scriptsize{$\pm$0.5} & 42.2 \scriptsize{$\pm$0.8} & 54.1 \scriptsize{$\pm$0.6} & & 67.6 \scriptsize{$\pm$0.5} & 63.6 \scriptsize{$\pm$0.6} & 71.9 \scriptsize{$\pm$0.5} & & 55.3 \scriptsize{$\pm$0.2} & 50.4 \scriptsize{$\pm$0.3} & 60.8 \scriptsize{$\pm$0.4} \\
                +Memoir (w/o retrieval)
                & & 59.6 & 50.1 & & 43.3 \scriptsize{$\pm$0.2} & 34.1 \scriptsize{$\pm$1.7} & 44.2 \scriptsize{$\pm$3.1} & & 63.0 \scriptsize{$\pm$0.3} & 52.9 \scriptsize{$\pm$0.3} & 61.0 \scriptsize{$\pm$0.5} & & 51.6 \scriptsize{$\pm$0.8} & 40.8 \scriptsize{$\pm$0.1} & 49.6 \scriptsize{$\pm$0.0} \\
                +Memoir (Ours)
                & & \bf 66.1 & \bf 61.3 & & \bf 50.2 \scriptsize{$\pm$0.3} & \bf 44.8 \scriptsize{$\pm$0.4} & \bf 56.2 \scriptsize{$\pm$0.6} & & \bf 69.8 \scriptsize{$\pm$0.2} & \bf 64.9 \scriptsize{$\pm$0.4} & \bf 73.3 \scriptsize{$\pm$0.2} & & \bf 57.7 \scriptsize{$\pm$0.1} & \bf 52.0 \scriptsize{$\pm$0.1} & \bf 61.9 \scriptsize{$\pm$0.4} \\
            \bottomrule
        \end{tabular}
        \begin{tablenotes}
            \footnotesize
            \item (w/o Retrieval): Variant pretrained and finetuned under identical conditions to Memoir, excluding only the explicit memory retrieval mechanism.
            \item[*] Results reproduced under aligned experimental conditions (episode ordering, training iterations, batch size, learning rate and dropout rate).
        \end{tablenotes}
    \end{threeparttable}
    }
    \label{tab:gsa_comprehensive}
\end{table*}

\subsection{Quantitative Analysis}
\subsubsection{Iterative Room-to-Room (IR2R)} \Cref{tab:ir2r_results} presents a comparison of Memoir against both traditional and memory-persistent methods on the IR2R benchmark. When applied to traditional VLN models, Memoir demonstrates substantial performance improvements: 11.1\% SPL enhancement for DUET-based implementations and 5.6\% for ScaleVLN-based implementations on unseen scenarios. These results prove that incorporating retrieved information from long-term memory serves as an effective prior for robust navigation decisions, even for models not originally designed for memory persistence. To disentangle the sources of improvement, we also explicitly analyze a \textit{w/o Retrieval} variant that benefits from training techniques like joint world model pretraining but lacks the active retrieval loop. Though this architectural baseline yields improvements on unseen environments, the complete Memoir framework further elevates performance, quantifying the substantial gain from the imagination-guided retrieval mechanism. When compared against memory-persistent approaches, Memoir significantly outperforms GR-DUET, achieving 5.4\% improvement in SPL on unseen scenarios (73.3\% versus 67.9\%) and 11.6\% improvement on seen scenarios. This superior performance validates our hypothesis that incorporating complete memory information introduces excessive noise that degrades navigation decisions and reduces flexibility in scenarios with limited experience availability. Our adaptive retrieval approach effectively addresses these limitations.

\begin{figure}
\centering
\includegraphics[scale=0.35]{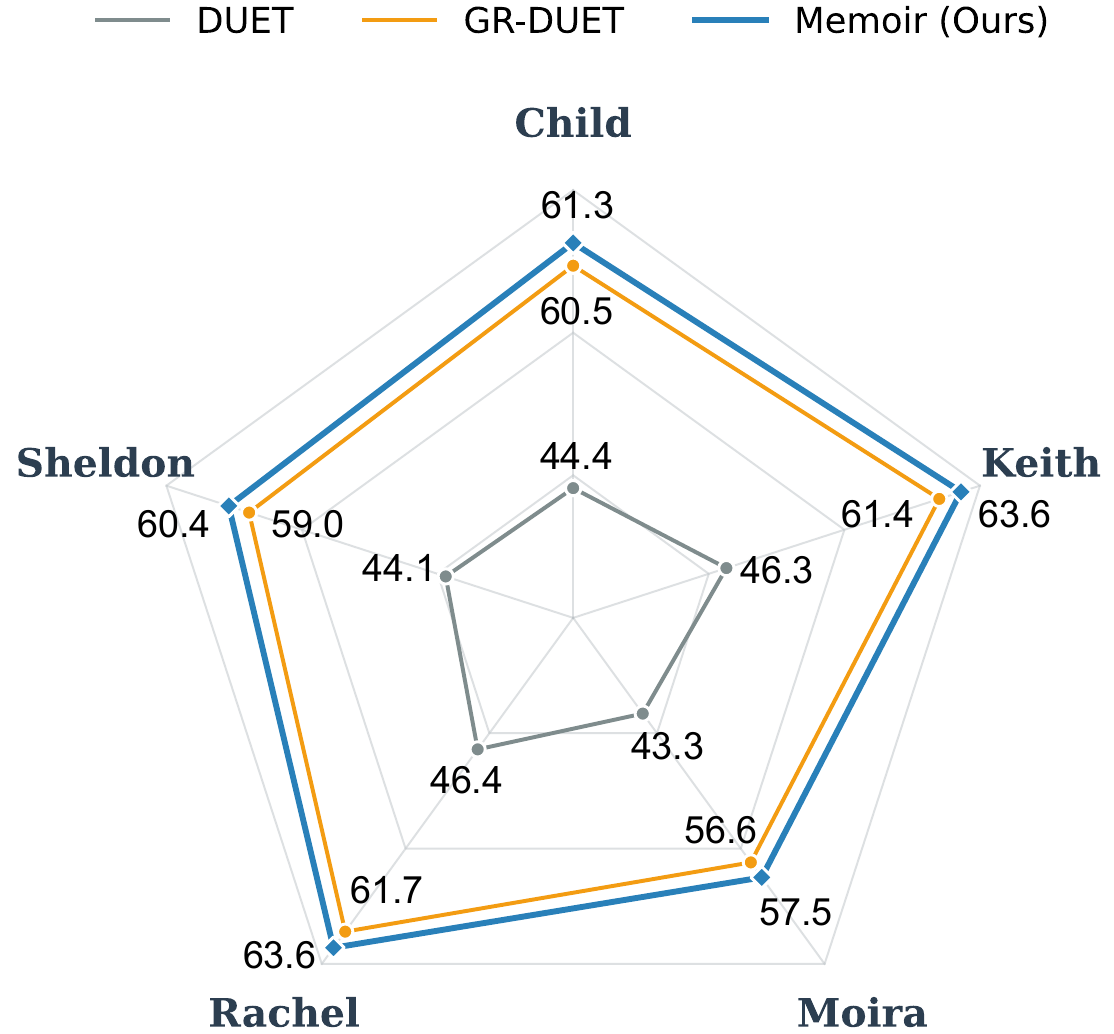}
\caption{Comparison of navigation performance (SPL) on various user instruction tasks from the GSA-R2R benchmark.}
\label{fig:radar}
\end{figure}

\noindent\textbf{Discussion.} While achieving exceptional performance on unseen scenarios, memory-persistent variants often exhibit reduced performance on seen scenarios compared to their traditional counterparts. For instance, DUET achieves 74.5\% SPL compared to GR-DUET's 55.1\% on seen environments, with our method experiencing approximately 3\% SPL degradation. This phenomenon stems from: (1) difference in tour lengths between validation splits, seen tours average only 6.4 episodes compared to 71.2 episodes in unseen tours, limiting accumulated experience; (2) regularization effects where long-term memory integration prevents overfitting to training environments by encouraging broader contextual reasoning rather than environmental detail memorization. Memoir substantially reduces this performance gap compared to GR-DUET, demonstrating more balanced memory utilization.

\subsubsection{General Scene Adaptation (GSA-R2R)} Tables~\ref{tab:gsa_comprehensive} summarize Memoir's performance across diverse scene adaptation scenarios. The user instructions taxonomy encompasses five tasks featuring distinct instruction styles. The performance comparison, presented in \Cref{tab:gsa_comprehensive} and \Cref{fig:radar}, demonstrates that our method consistently outperforms the GR-DUET baseline by 2.3\% in SR and 2.5\% in SPL on average. The scene instructions taxonomy and basic instructions taxonomy evaluate performance across different environmental characteristics and instruction expressions. Memoir consistently outperforms both adaptation-based and memory-based methods, achieving an average 2.4\% SR increase and 1.6\% SPL improvement compared to GR-DUET across eight distinct testing scenarios with aligned experimental configurations. The improvements demonstrate that hybrid memory provides critical context absent in traditional approaches: by accessing past episodes where agents successfully processed similar expressions and executed corresponding actions, Memoir learns from historical patterns that GR-DUET's observation memory cannot capture.

\noindent\textbf{Discussion.} Memoir consistently outperforms GR-DUET, though with smaller margins than on IR2R. This reduced improvement stems from memory density differences, with GSA-R2R accumulating 600 episodes on average, increasing the topological completeness for GR-DUET.

\begin{figure*}[t]
\centering
\includegraphics[scale=0.55]{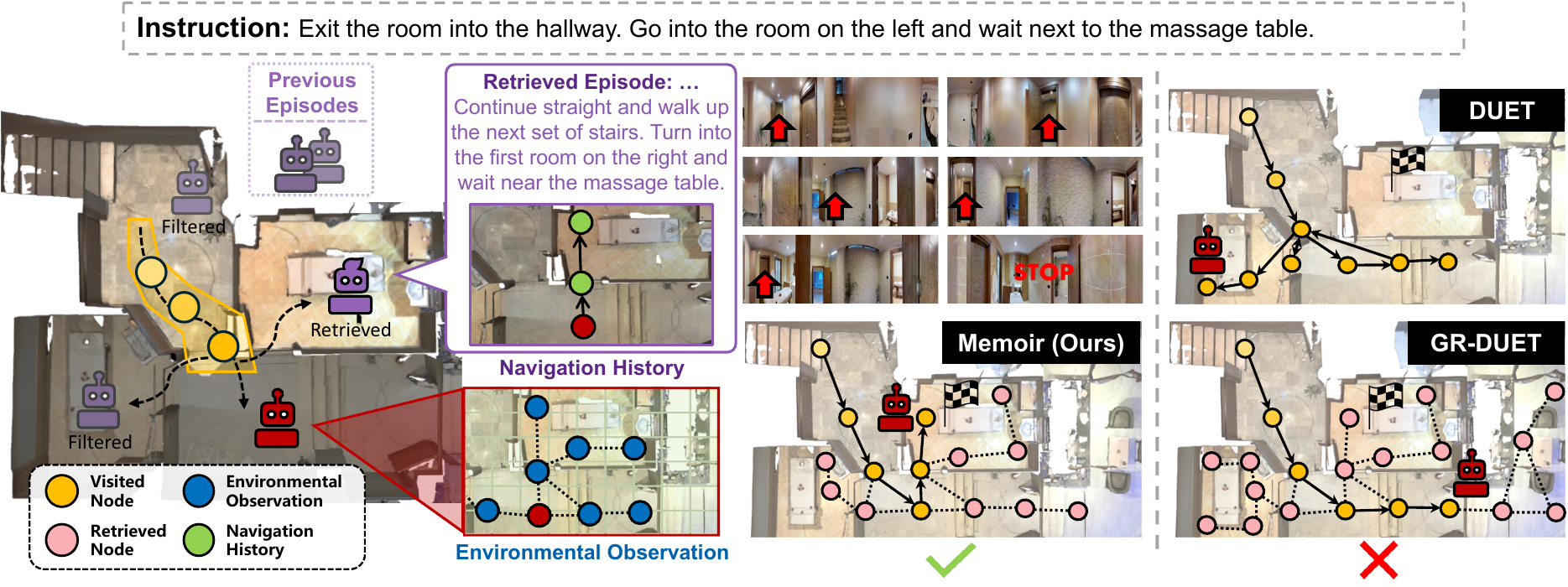}
\caption{Visualization of Memoir's memory retrieval from environmental observation bank and navigation history bank as well as the panoramic trajectory visualization. We compare the navigation result between DUET, GR-DUET and ours. The goal location is indicated by checkered flag.}
\label{fig:case_study}
\end{figure*}

\begin{table}[t]
    \renewcommand{\arraystretch}{1.15}
    \setlength{\aboverulesep}{0pt}
    \setlength{\belowrulesep}{0pt}
    \centering
    \caption{
	    Computational efficiency comparison (batch size = 4).
	}
	\resizebox{\linewidth}{!}{
    \begin{threeparttable}
		\begin{tabular}{lcccccc}
			\toprule
            & & \multicolumn{2}{c}{\textbf{Training}} & & \multicolumn{2}{c}{\textbf{Inference}}
            \\
			\cmidrule{3-4} \cmidrule{6-7}
			Methods
			&
			& \textbf{Memory}~$\downarrow$
			& \textbf{Latency}~$\downarrow$
			&
			& \textbf{Memory}~$\downarrow$
			& \textbf{Latency}~$\downarrow$
			\\
			\midrule
			\midrule
                DUET \cite{2022duet}
                & & 7.2 GB & 0.15s & & 2.2 GB & 0.13s \\
                GR-DUET \cite{2025gsa}
                & & 29.4 GB & 4.39s & & 9.9 GB & 0.25s \\
                Memoir (Ours)
                & & \bf 13.1 GB (-55\%) & \bf 0.53s (-88\%) & & \bf 2.6 GB (-74\%) & 0.31s (+28\%) \\
			\bottomrule
		\end{tabular}
    \end{threeparttable}
    }
	\label{tab:computational_efficiency}
\end{table}

\begin{figure}[t]
    \centering
    \includegraphics[scale=0.31]{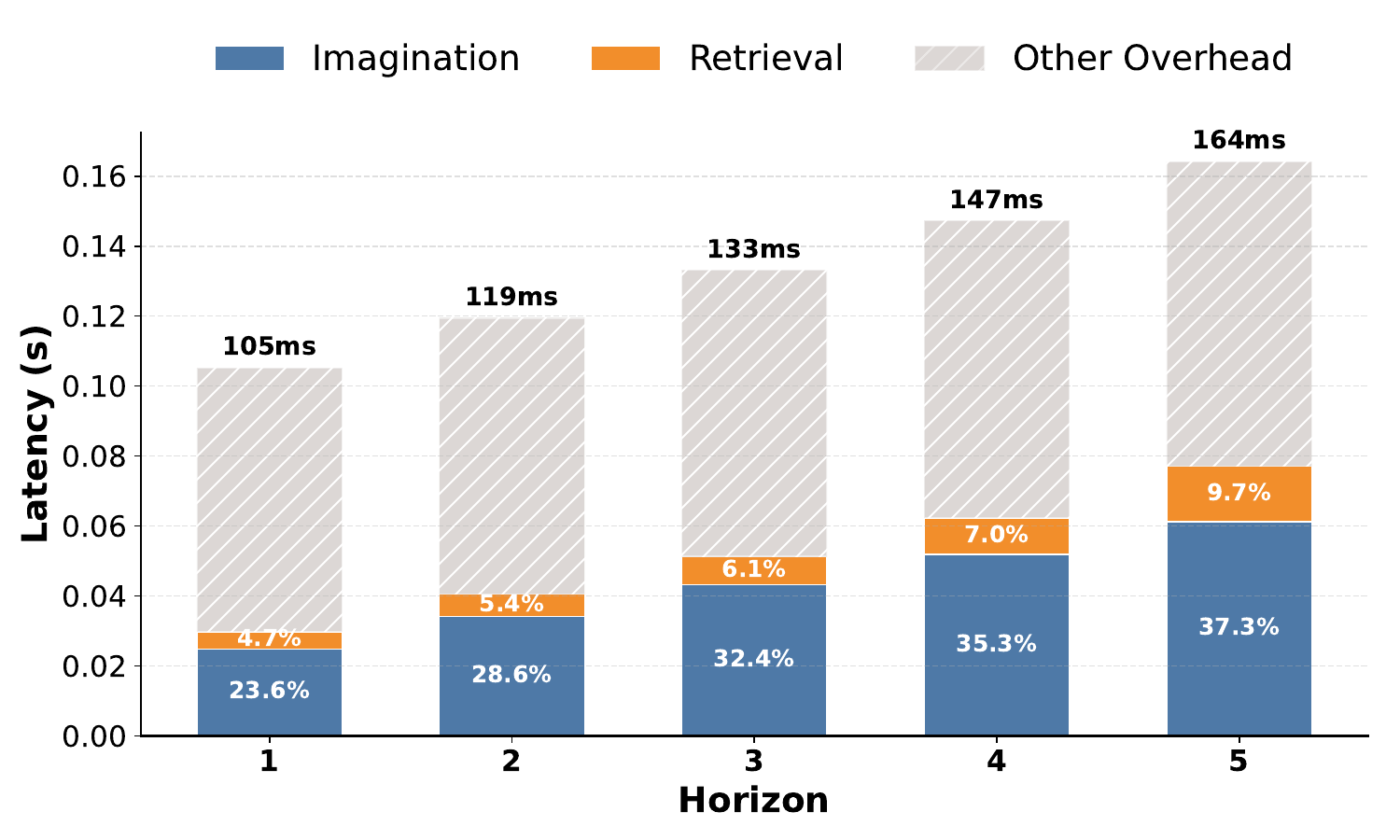}
    \caption{Averaged inference latency breakdown in navigation as the imagination horizon ($D$) increases. (batch size=1)}
    \label{fig:latency_breakdown}
\end{figure}

\subsection{Qualitative Analysis}
\Cref{fig:case_study} demonstrates memory retrieval effectiveness in challenging scenarios where DUET and GR-DUET fail. Given the task of locating a ``massage table" with two potential candidates barely visible from the hallway, the DUET agent incorrectly approaches the wrong target without observing the actual target, while the GR-DUET agent becomes confused among numerous candidate locations and produces incorrect decisions. Our model succeeds through the combination of observation retrieval, which identifies promising paths toward relevant locations while controlling redundancy, and history retrieval, which matches similar past episodes targeting ``massage room" objectives, prompting the agent to the destination.

\begin{figure}[t]
    \centering
    \subfloat[Cum. SR v.s. Episode Count.]{
        \includegraphics[width=0.45\linewidth]{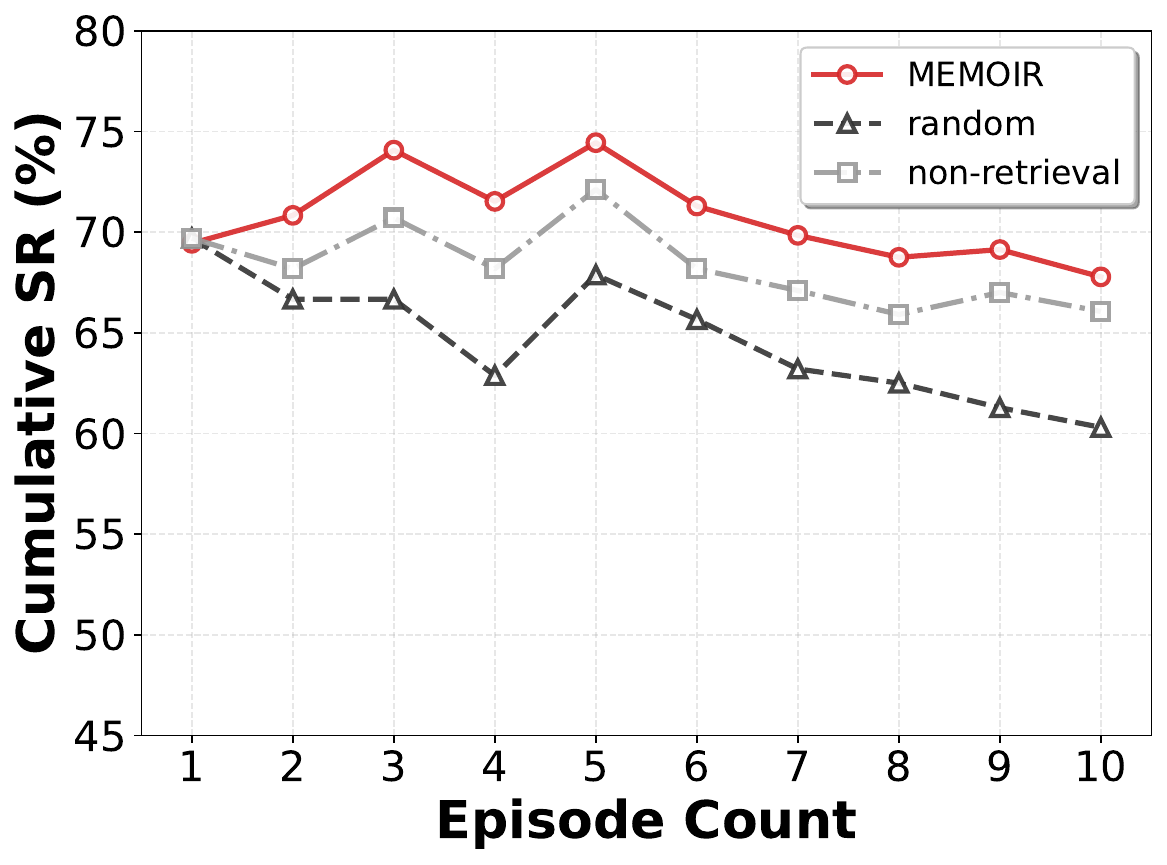}
        \label{fig:episode_sr}
    }
    \hfill
    \subfloat[Cum. SPL v.s. Episode Count.]{
        \includegraphics[width=0.45\linewidth]{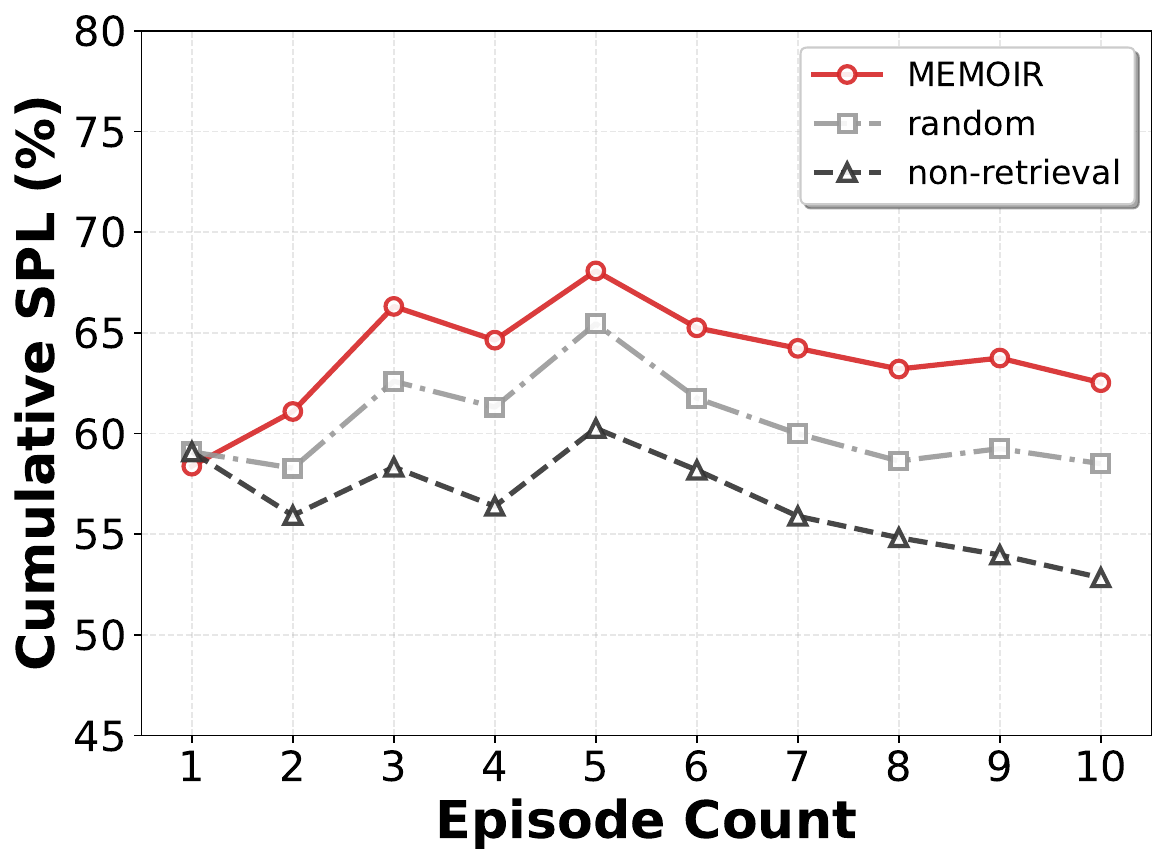}
        \label{fig:episode_spl}
    }
    
    \vspace{-1.0em}

    \subfloat[Cum. SR v.s. Tour Progress.]{
        \includegraphics[width=0.45\linewidth]{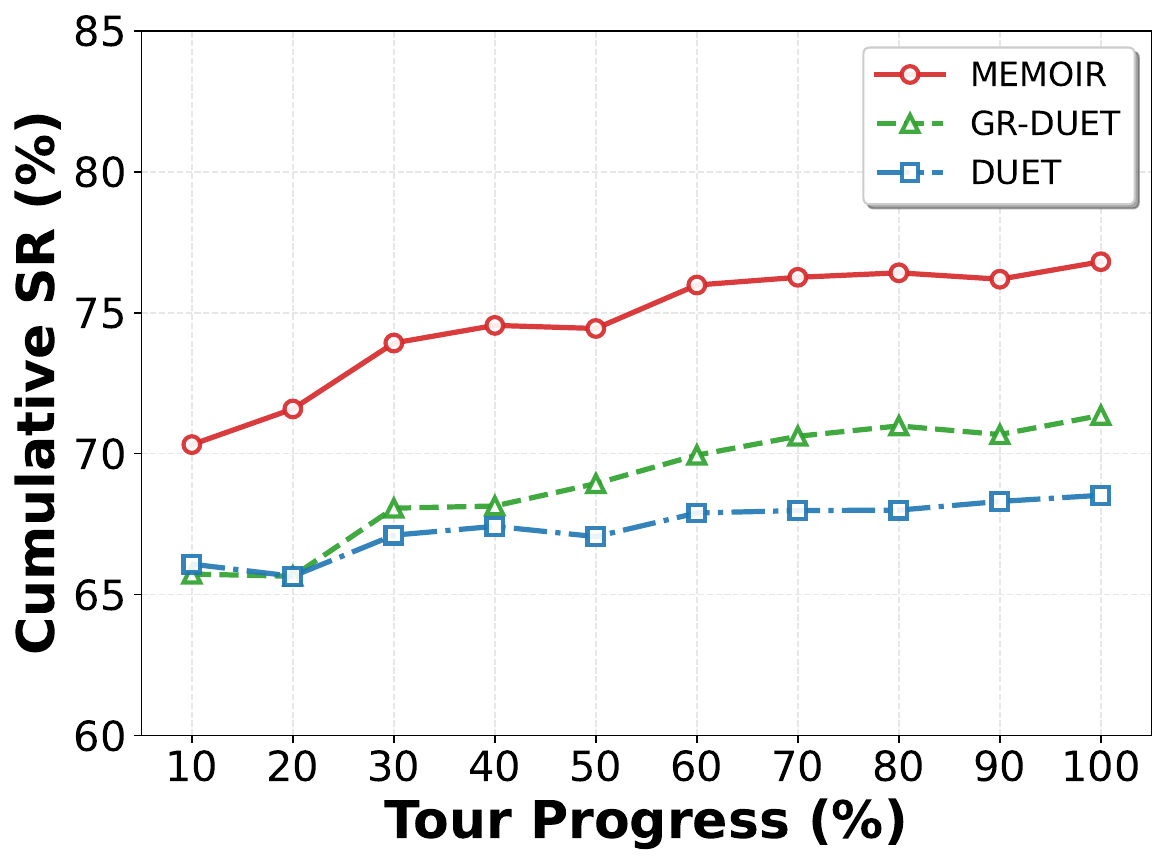}
        \label{fig:tour_sr}
    }
    \hfill
    \subfloat[Cum. SPL v.s. Tour Progress.]{
        \includegraphics[width=0.45\linewidth]{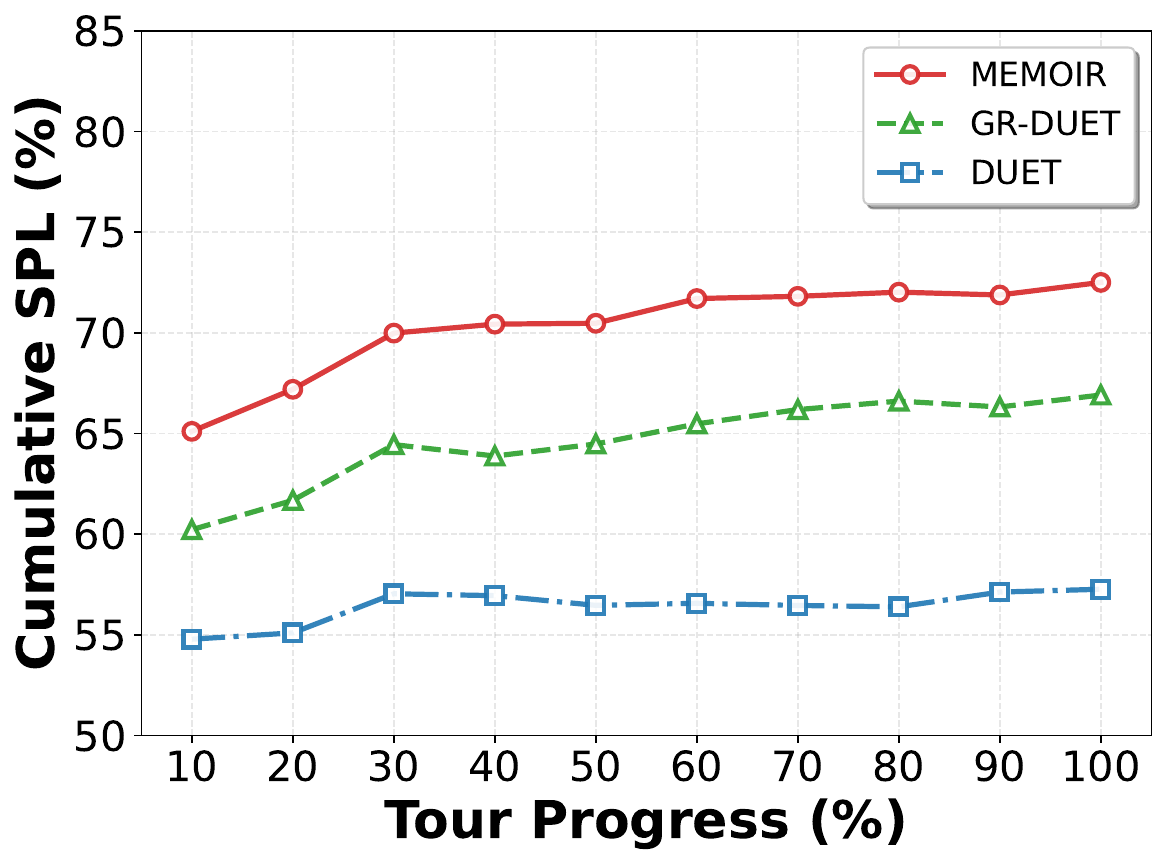}
        \label{fig:tour_spl}
    }
    
    \caption{Cumulative performance scaling across tour progression on IR2R.}
    \label{fig:tour_spl_combined}
\end{figure}

\subsection{Ablation Studies \& Analyses}

\noindent\textbf{Computational Efficiency.}
\Cref{tab:computational_efficiency} demonstrates Memoir's computational advantages over the memory-persistent baseline. While DUET operates with minimal memory overhead, GR-DUET's complete memory retention strategy dramatically increases resource requirements (29.4GB training, 9.9GB inference) due to processing all accumulated observations simultaneously. Our retrieval mechanism achieves substantial efficiency gains: 55\% reduction in training memory and 88\% reduction in training latency, representing an 8.3× speedup. During inference, memory usage decreases by 74\%, approaching DUET's efficiency while maintaining memory-persistent capabilities. To further investigate the inference overhead, we analyze the latency scalability in \Cref{fig:latency_breakdown}. The slight increase in total inference latency (0.31s vs 0.25s) is primarily driven by the imagination process, which scales linearly with the lookahead horizon ($D$). Crucially, the retrieval latency remains remarkably low (approximately 16ms even at $D=5$) and exhibits sub-linear growth relative to the total time. With a total latency consistently under 170ms per step across horizons, Memoir establishes itself as an approach to achieve both SOTA performance and practical real-time feasibility for resource-constrained deployment.

\begin{table*}[t]
    \renewcommand{\arraystretch}{1.15}
    \setlength{\aboverulesep}{0pt}
    \setlength{\belowrulesep}{0pt}
    \setlength{\tabcolsep}{5.5pt}
    \centering
    \caption{
        Ablation of components for memory retrieval.
    }
    \resizebox{0.85\linewidth}{!}{
    \begin{threeparttable}
        \begin{tabular}{l cc c cc c ccccccccc}
            \toprule
            & \textbf{Observation} & & \textbf{History} & & \multicolumn{9}{c}{\textbf{IR2R Val Unseen}}
            \\
            \cmidrule{2-2} \cmidrule{4-4} \cmidrule{6-14}
            & Strategy ($\mathcal{M}_o$) & & Strategy ($\mathcal{M}_h$) & & \textbf{\texttt{TL}}~$\downarrow$
            & \textbf{\texttt{NE}}~$\downarrow$
            & \textbf{\texttt{SR}}~$\uparrow$
            & \textbf{\texttt{SPL}}~$\uparrow$
            & \textbf{\texttt{nDTW}}~$\uparrow$
            & \textbf{\texttt{OR}}~$\uparrow$
            & \textbf{\texttt{OA}}~$\uparrow$
            & \textbf{\texttt{HR}}~$\uparrow$
            & \textbf{\texttt{HA}}~$\uparrow$
            \\
            \midrule
            \midrule

            \rowcolor{gray!15}\multicolumn{14}{c}{\emph{The upper-bound of long-term memory retrieval}}\\
            & \textsc{Oracle} & & \textsc{Oracle} & & 9.77 & 0.51 & 95.44 & 93.40 & 93.68 & 100 & 100 & 100 & 100 \\
            
            \midrule

            & \textsc{None} & & \textsc{None} & & 12.24 & 2.81 & 72.33 & 63.97 & 70.35 & 0.0 & 0.0 & 0.0 & 0.0 \\
            & \textsc{Full} & & \textsc{Full} & & 10.44 & 2.86 & 74.67 & 69.98 & 76.29 & 100 & 9.81 & 100 & 19.33 \\
            & \textsc{Random} & & \textsc{Random} & & 10.97 & 2.76 & 75.82 & 70.34 & 76.03 & 59.05 & 21.31 & 36.65 & 22.11 \\

            & \textsc{Random} & & \textsc{Imagination} & & 10.80 & 2.61 & 76.63 & 71.03 & 76.98 & 58.83 & 21.72 & 98.36 & 22.40 \\
            & \textsc{Imagination} & & \textsc{Random} & & 10.77 & 2.58 & 76.63 & 71.70 & 78.08 & 97.05 & 23.33 & 36.24 & 21.94 \\
            & \textsc{Imagination} & & \textsc{Instruction} & & 10.28 & 2.59 & 77.01 & 72.82 & 79.43 & 97.77 & 22.04 & 97.92 & 21.82 \\
            & \textsc{Imagination} & & \textsc{State} & & 10.11 & 2.67 & 76.54 & 72.85 & 79.16 & 97.13 & 23.19 & 81.86 & 23.49 \\

            & \textsc{Imagination} & & \textsc{Imagination} & & 10.32 & \textbf{2.53} & \textbf{78.03} & \textbf{73.46} & \textbf{79.46} & 96.49 & \textbf{24.58} & 96.52 & \textbf{24.21} \\
            \bottomrule
        \end{tabular}
        \begin{tablenotes}
            \footnotesize
            \item \textsc{Imagination}: retrieve via imagination; \textsc{Random}: random sampling; \textsc{Full}: full incorporation; \textsc{Oracle}: optimal retrieval; \textsc{Instruction}: retrieve via instruction-similarity; \textsc{State}: retrieve via state-similarity.
        \end{tablenotes}
    \end{threeparttable}
    }
    \label{tab:ablation}
\end{table*}

\begin{table}[t]
    \centering
    \renewcommand{\arraystretch}{1.0}
    \caption{
        Ablation of World Model Variants.
    }
    \label{tab:world_model}

    \begin{minipage}{\linewidth}
        \centering
        \textbf{(a) Impact of Backbone Architecture (Fixed $D=5$)}
        \resizebox{\linewidth}{!}{
            \begin{tabular}{l c cccccc}
                \toprule
                \textbf{Backbone} & \textbf{Overshoot} & \textbf{SR}~$\uparrow$ & \textbf{SPL}~$\uparrow$ & \textbf{OR}~$\uparrow$ & \textbf{OA}~$\uparrow$ & \textbf{HR}~$\uparrow$ & \textbf{HA}~$\uparrow$ \\
                \midrule
                GRU &  & 76.12 & 71.48 & \textbf{97.54} & 16.90 & 30.21 & 30.78 \\
                GRU & \checkmark & 76.71 & 72.30 & 96.21 & 18.60 & \textbf{98.24} & 21.65 \\
                \midrule
                Transformer &  & 77.14 & 72.11 & 96.44 & 17.86 & 28.58 & \textbf{32.32} \\
                Transformer & \checkmark & \textbf{78.03} & \textbf{73.46} & 96.49 & \textbf{24.58} & 96.52 & 24.21 \\
                \bottomrule
            \end{tabular}
        }
    \end{minipage}
    
    \vspace{3pt}

    \begin{minipage}{\linewidth}
        \centering
        \textbf{(b) Impact of Imagination Horizon (Transformer-based)}
        \resizebox{\linewidth}{!}{
            \begin{tabular}{c c c ccccc}
                \toprule
                \textbf{Overshoot} & \textbf{Horizon ($D$)} & \textbf{SR}~$\uparrow$ & \textbf{SPL}~$\uparrow$ & \textbf{OR}~$\uparrow$ & \textbf{OA}~$\uparrow$ & \textbf{HR}~$\uparrow$ & \textbf{HA}~$\uparrow$ \\
                \midrule
                 & 1 & 73.44 & 67.24 & 31.25 & 40.33 & 26.26 & 30.74 \\
                 & 3 & 75.18 & 70.21 & 76.71 & 26.05 & 29.11 & 32.29 \\
                 & 5 & 77.14 & 72.11 & 96.44 & 17.86 & 28.58 & \textbf{32.32} \\
                \midrule
                \checkmark & 1 & 73.09 & 67.26 & 31.15 & \textbf{43.22} & 59.43 & 29.24 \\
                \checkmark & 3 & 76.63 & 71.99 & 77.04 & 28.48 & 93.40 & 25.06 \\
                \checkmark & \textbf{5} & \textbf{78.03} & \textbf{73.46} & \textbf{96.49} & 24.58 & \textbf{96.52} & 24.21 \\
                \bottomrule
            \end{tabular}
        }
    \end{minipage}
\end{table}

\noindent\textbf{Performance Scaling.}
We analyze performance scaling through both micro-level accumulation (\Cref{fig:episode_sr} and \Cref{fig:episode_spl}) and macro-level tour progression (\Cref{fig:tour_sr} and \Cref{fig:tour_spl}) .
At the micro-level, all variants exhibit identical initial performance, confirming architectural parity; however, Memoir immediately diverges with a steep upward trajectory above other variants. Crucially, although random retrieval approximates our SR through stochastic coverage, both random and non-retrieval variants consistently underperform in SPL with a widening gap. This confirms that our imagination-guided mechanism optimizes navigation efficiency, rather than merely benefiting goal discovery. On the macro-level, Memoir maintains a robust and constant lead over GR-DUET throughout the tour, demonstrating consistent long-term adaptability compared to the baseline's suboptimal performance.

\begin{table}[t]
    \centering
    \caption{
        Ablation of navigation history integration.
    }
    \resizebox{0.9\linewidth}{!}{
        \begin{tabular}{cccccc}
            \toprule
            & \multicolumn{4}{c}{\textbf{IR2R Val Unseen}} \\
            \textsc{Hist Encoder} & \textsc{Embed Type}
            & \textbf{\texttt{SR}}~$\uparrow$
            & \textbf{\texttt{SPL}}~$\uparrow$
            & \textbf{\texttt{nDTW}}~$\uparrow$
            & \textbf{\texttt{t-nDTW}}~$\uparrow$
            \\
            \midrule
            & VP + State & 76.59 & 72.35 & 78.60 & 65.57 \\
            \checkmark & VP & 76.54 & 71.48 & 78.24 & 65.66 \\
            \checkmark & State & 77.35 & 72.83 & 78.70 & 66.37 \\
            \checkmark & VP + State & \textbf{78.03} & \textbf{73.46} & \textbf{79.46} & \textbf{66.44} \\
            \bottomrule
        \end{tabular}}
    \label{tab:history_representation}
\end{table}

\begin{table}[t]
    \centering
    \caption{
        Ablation of expert policies.
    }
    \resizebox{0.9\linewidth}{!}{
        \begin{tabular}{lccccccc}
            \toprule
            & \multicolumn{3}{c}{\textbf{IR2R Val Unseen}} && \multicolumn{3}{c}{\textbf{GSA Test-R-Basic}} \\
            \cmidrule{2-4} \cmidrule{6-8}
            \textsc{Expert Policy}
            & \textbf{\texttt{SR}}~$\uparrow$
            & \textbf{\texttt{SPL}}~$\uparrow$
            & \textbf{\texttt{nDTW}}~$\uparrow$
            &
            & \textbf{\texttt{SR}}~$\uparrow$
            & \textbf{\texttt{SPL}}~$\uparrow$
            & \textbf{\texttt{nDTW}}~$\uparrow$
            \\
            \midrule
            SPL & 76.20 & 71.71 & 78.03 && 67.34 & 62.22 & 71.30 \\
            +random sample & \bf 78.03 & \bf 73.46 & \bf 79.46 && \bf 69.64 & \bf 64.91 & \bf 73.29 \\
            \bottomrule
        \end{tabular}}
    \label{tab:expert}
\end{table}

\noindent\textbf{Memory Retrieval.}
\Cref{tab:ablation} presents results validating the effectiveness of memory components. The ``oracle" variant directly incorporates memories leading to target locations, simulating ideal world model behavior with perfect retrieval capabilities, achieving 93.40\% SPL on unseen environments. This highlights the necessity of retrieval and serves as a performance upper bound. The variant with both observation and history components disabled yields the lowest performance, followed by complete long-term memory incorporation. Random memory selection enhances navigation performance by 0.36\% SPL. Instruction-based retrieval achieves high recall (97.92\% HR) by matching global semantics but suffers from temporal misalignment, retrieving entire trajectories without localizing the specific segment relevant to current progress (21.82\% HA). State-based retrieval improves precision (23.49\% HA) but suffers from path dependency; because differences in past paths prevent the retrieval of spatially relevant experiences even if the future goal is identical (81.86\% HR). In contrast, our imagination-based retrieval achieves optimal navigation performance (73.46\% SPL) and retrieval accuracy (24.58\% OA, 24.21\% HA). By querying with the imagined latents, Memoir grounds retrieval in navigation intent, overcoming the noise of static instruction matching and the rigidity of historical state matching. The substantial gap relative to the ideal world model reveals current challenges in the world model's ability to capture environmental dynamics accurately, suggesting benefits from future data scaling.

\noindent\textbf{World Model.}
\Cref{tab:world_model}(a) evaluates world model variants comparing GRU and Transformer architectures. Transformer variant outperforms GRU variant by 1.16\% SPL, 7.68\% observation retrieval accuracy (OA), and 2.56\% history retrieval accuracy (HA) under 5-step overshooting distance. As illustrated in \Cref{tab:world_model}(b), retrieval effectiveness generally increases as the exploration horizon broadens, as recall improves for observations (OR) and histories (HR), enabling more informed navigation decisions. However, retrieval accuracy decreases due to increasing candidates in topological graphs with greater distances. The overshooting objective significantly enhances OA and HR, contributing to robust navigation performance (+1.36\% SPL). These results validate that effective retrieval requires both powerful predictive models (Transformer over GRU) and grounded training (overshooting) to balance exploration breadth with query precision.

\begin{table}[t]
    \centering
    \caption{
        Ablation of world model pretraining.
    }
    \resizebox{0.8\linewidth}{!}{
        \begin{tabular}{cccccccc}
            \toprule
            & \multicolumn{3}{c}{\textbf{IR2R Val Unseen}} && \multicolumn{3}{c}{\textbf{GSA Test-R-Basic}} \\
            \cmidrule{2-4} \cmidrule{6-8}
            \textsc{Pretrain}
            & \textbf{\texttt{SR}}~$\uparrow$
            & \textbf{\texttt{SPL}}~$\uparrow$
            & \textbf{\texttt{nDTW}}~$\uparrow$
            &
            & \textbf{\texttt{SR}}~$\uparrow$
            & \textbf{\texttt{SPL}}~$\uparrow$
            & \textbf{\texttt{nDTW}}~$\uparrow$
            \\
            \midrule
             & 74.93 & 71.55 & 78.12 && 64.62 & 61.52 & 71.30 \\
            \checkmark & \bf 78.03 & \bf 73.46 & \bf 79.46 && \bf 69.64 & \bf 64.91 & \bf 73.29 \\
            \bottomrule
        \end{tabular}}
    \label{tab:pretrain}
\end{table}

\begin{table}[t]
    \centering
    \caption{
        Ablation of observation completion.
    }
    \resizebox{0.85\linewidth}{!}{
        \begin{tabular}{cccccccc}
            \toprule
            & \multicolumn{3}{c}{\textbf{IR2R Val Unseen}} && \multicolumn{3}{c}{\textbf{GSA Test-R-Basic}} \\
            \cmidrule{2-4} \cmidrule{6-8}
            \textsc{Neighbor obs}
            & \textbf{\texttt{SR}}~$\uparrow$
            & \textbf{\texttt{SPL}}~$\uparrow$
            & \textbf{\texttt{nDTW}}~$\uparrow$
            &
            & \textbf{\texttt{SR}}~$\uparrow$
            & \textbf{\texttt{SPL}}~$\uparrow$
            & \textbf{\texttt{nDTW}}~$\uparrow$
            \\
            \midrule
            Partial & 77.05 & 72.15 & 78.22 && 68.35 & 63.90 & 72.89 \\
            Completion & \bf 78.03 & \bf 73.46 & \bf 79.46 && \bf 69.64 & \bf 64.91 & \bf 73.29 \\
            \bottomrule
        \end{tabular}}
    \label{tab:observation_completion}
\end{table}

\begin{table}[t]
    \centering
    \caption{
        Ablation of neighborhood incorporation.
    }
    \resizebox{0.85\linewidth}{!}{
        \begin{tabular}{cccccccc}
            \toprule
            & \multicolumn{3}{c}{\textbf{IR2R Val Unseen}} && \multicolumn{3}{c}{\textbf{GSA Test-R-Basic}} \\
            \cmidrule{2-4} \cmidrule{6-8}
            \textsc{Retrieve}
            & \textbf{\texttt{SR}}~$\uparrow$
            & \textbf{\texttt{SPL}}~$\uparrow$
            & \textbf{\texttt{nDTW}}~$\uparrow$
            &
            & \textbf{\texttt{SR}}~$\uparrow$
            & \textbf{\texttt{SPL}}~$\uparrow$
            & \textbf{\texttt{nDTW}}~$\uparrow$
            \\
            \midrule
            VP only & 76.46 & 70.91 & 77.45 && 67.44 & 63.02 & 71.83 \\
            VP + neighbors & \bf 78.03 & \bf 73.46 & \bf 79.46 && \bf 69.64 & \bf 64.91 & \bf 73.29 \\
            \bottomrule
        \end{tabular}}
    \label{tab:neighbors}
\end{table}

\begin{figure}[t]
    \centering
    \subfloat[Param Search (Obs.)]{
        \includegraphics[width=0.46\linewidth]{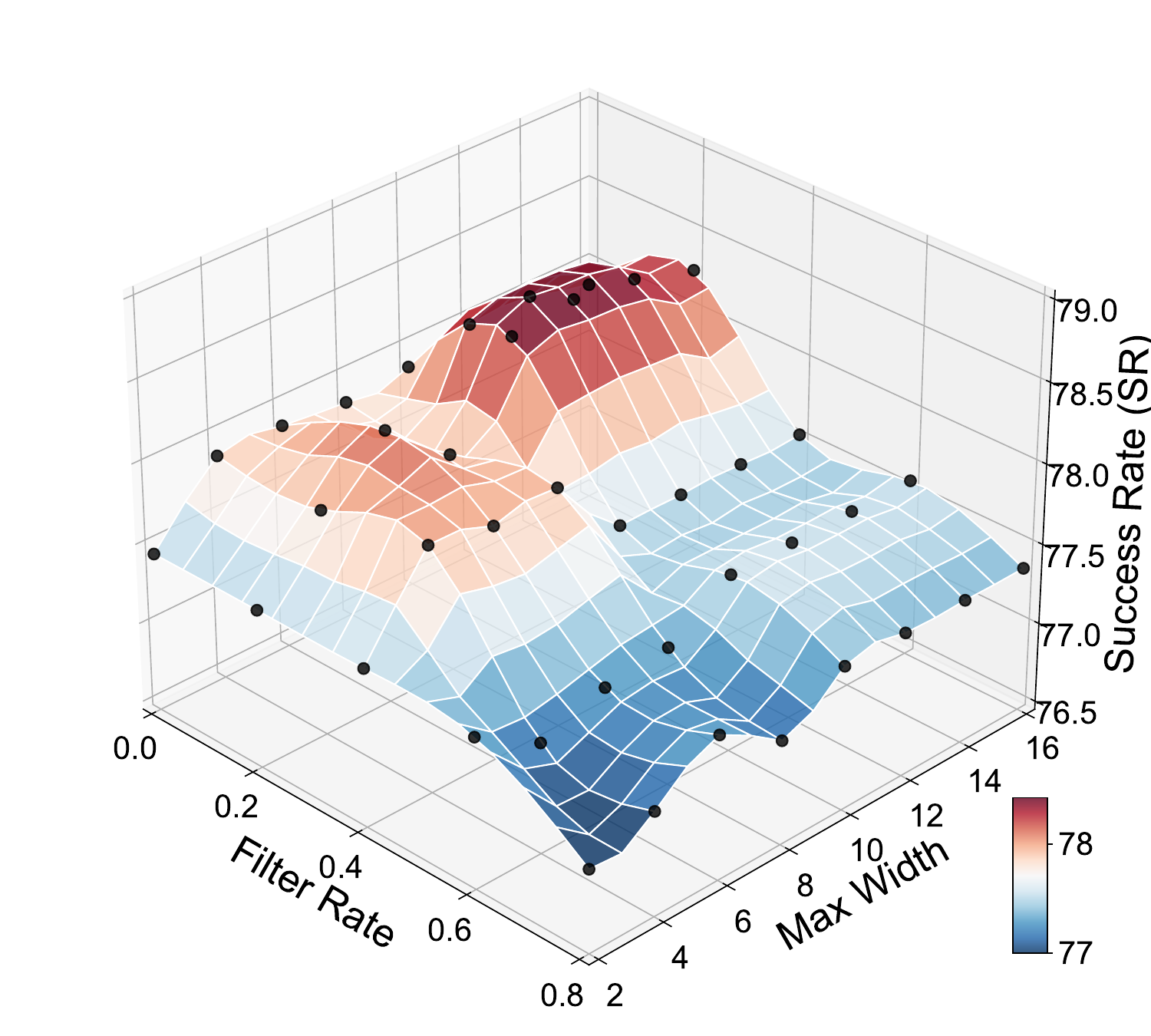}
        \label{fig:obs_memory}
    }
    \hfill
    \subfloat[Param Search (Hist.)]{
        \includegraphics[width=0.42\linewidth]{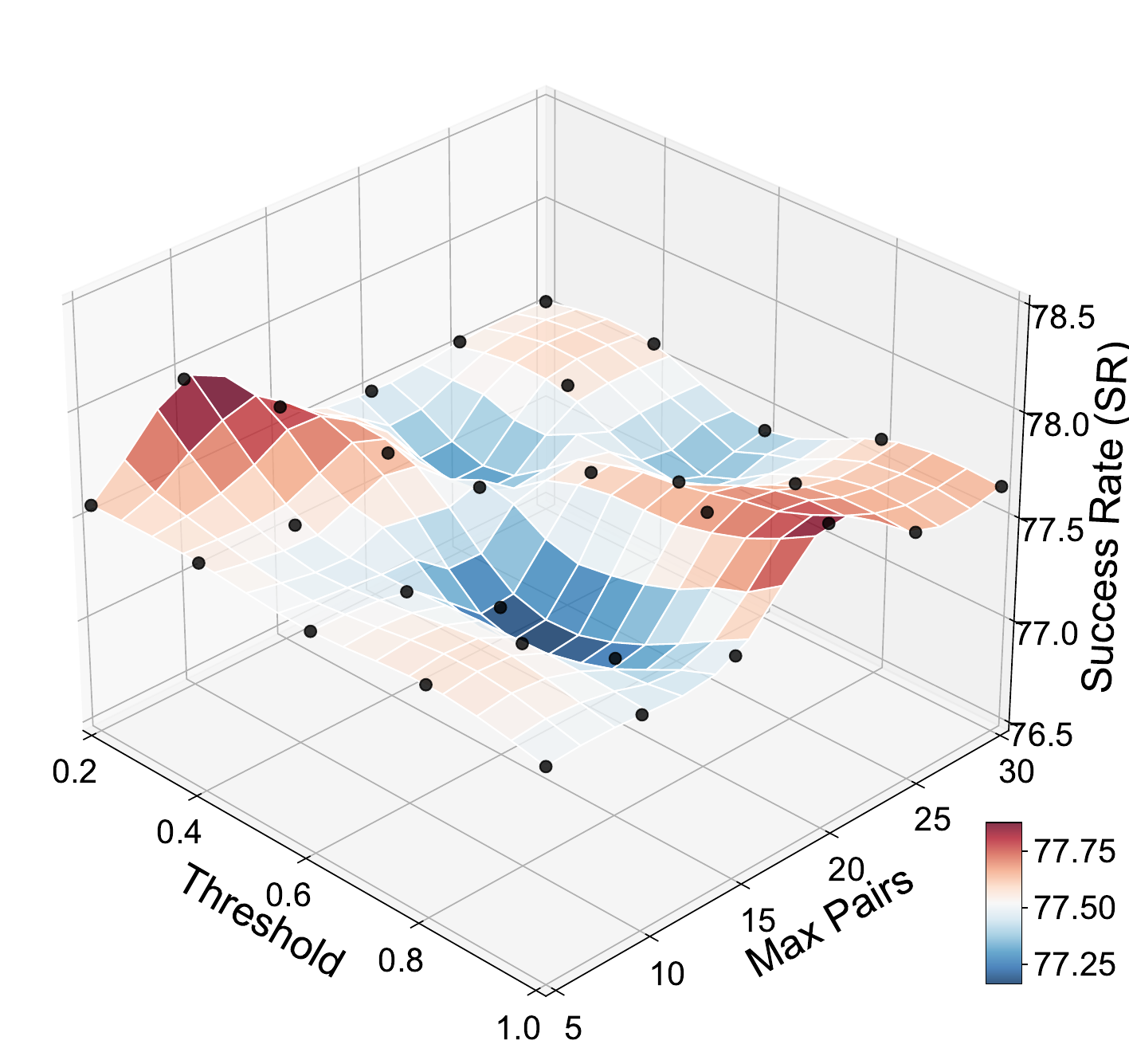}
        \label{fig:nav_memory}
    }
    \caption{Study of hyper-parameters of retrieval on IR2R val-unseen. Left: Environmental observation retrieval. Right: Navigation history retrieval.}
    \label{fig:param_search}
\end{figure}

\noindent\textbf{History Integration.}
\Cref{tab:history_representation} compares strategies for incorporating retrieved histories. Concatenating viewpoint features with state features in the dedicated encoder yields optimal performance, achieving 1.98\% higher SPL than viewpoint features alone and 0.63\% higher than state features alone. This indicates that state features carry crucial pattern information, while still benefiting from observation representation enhancement. Without incorporating history encoder, where historical representation is concatenated with coarse-scale encoder inputs, performance decreases by 1.11\% SPL, demonstrating the necessity of separating duties across three distinct encoders.

\noindent\textbf{Expert Policy Strategy.} \Cref{tab:expert} compares training strategies when memory retrieval dynamically expands the available action space beyond immediate neighbors. Random sampling among multiple optimal paths during training (achieving 73.46\% SPL) outperforms deterministic SPL-based expert selection (71.71\% SPL) by providing better policy regularization and robustness to navigation choices.

\noindent\textbf{World Model Pretraining.} \Cref{tab:pretrain} demonstrates the importance of proper world model initialization. Pretraining the world model components on navigation trajectories before joint training improves performance by 1.91\% SPL on IR2R and 3.39\% SPL on GSA-R2R, indicating that randomly initialized world models provide poor retrieval signals.

\begin{figure*}[t]
\centering
\includegraphics[scale=0.44]{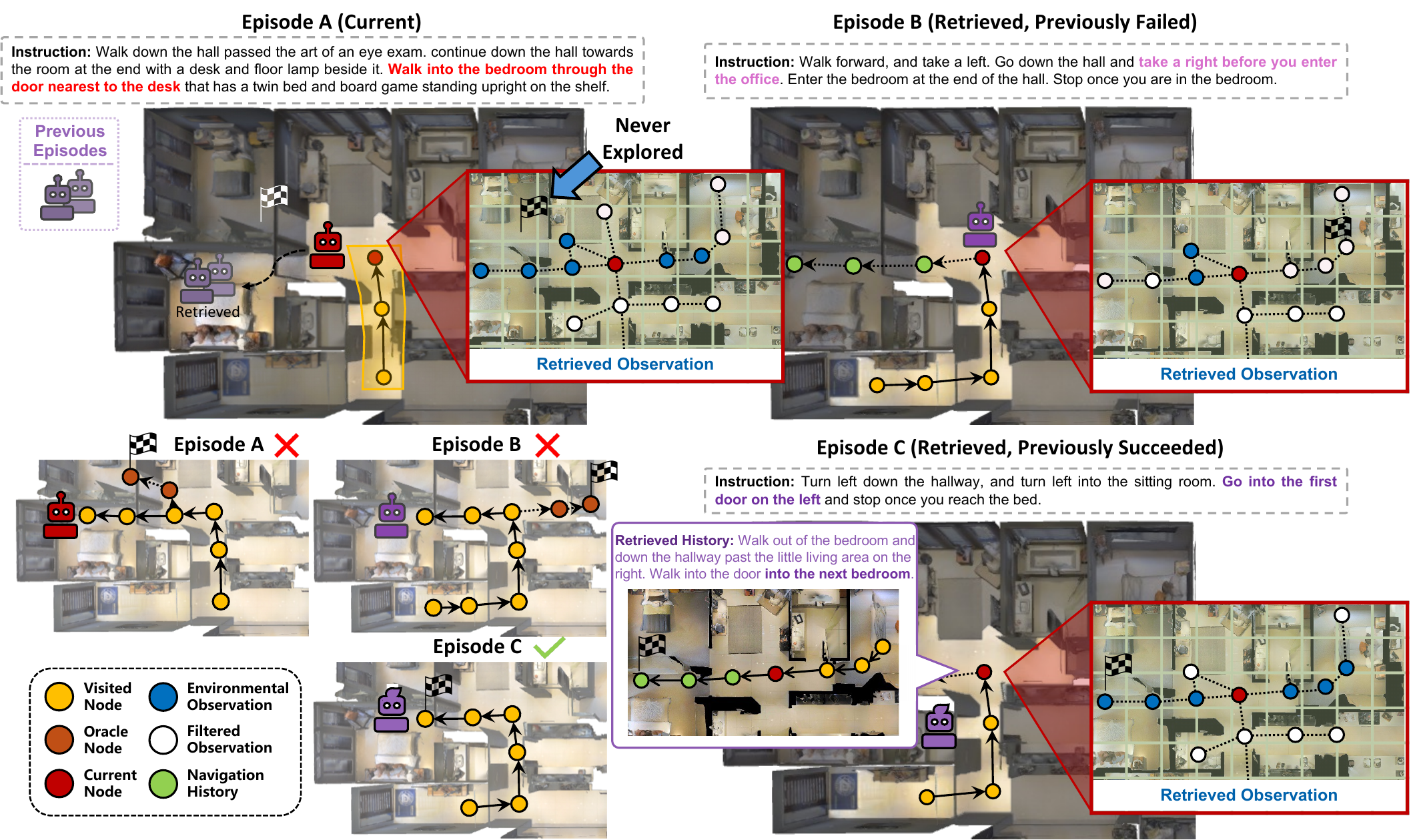}
\caption{Visualization of failure modes in imagination-guided memory retrieval. Episode A (current) retrieves experiences from Episodes B (previously failed) and C (previously succeeded) at a critical decision point. Key instruction differences are highlighted in bold.}
\label{fig:failure_cases}
\end{figure*}

\noindent\textbf{Observation Completion.} \Cref{tab:observation_completion} demonstrates that completing partial observations at non-retrieved viewpoints using stored features from $\mathcal{M}_o$ significantly enhances environmental understanding. When a viewpoint in the episodic graph lacks complete visual information, retrieving its stored panoramic feature enables more informed decision-making.

\noindent\textbf{Neighbor Incorporation.} \Cref{tab:neighbors} studies the retrieval strategy that incorporates adjacent viewpoints of retrieved nodes during observation retrieval. Including immediate neighbors provides richer spatial context about connectivity and surrounding environment, enabling the coarse-scale encoder to make better-informed planning decisions. This approach improves SPL by 2.55\% and 1.89\% on respective benchmark.

\noindent\textbf{Parameter Study.}
\Cref{fig:param_search} analyzes the impact of key retrieval hyperparameters. For observation retrieval, SR improves with reduced filter rates and increased search width, peaking at $\rho_o=0.2$ and $W=12$ before degrading as excessive context introduces noise. This indicates incorporating a broader range of viewpoint observations facilitates more informed navigation decisions. For navigation history retrieval, the model prioritizes precision over recall, achieving optimal performance at $\theta_h=0.2$ with $P=10$. A secondary optimum occurs at threshold $\theta_h=1.0$ and max patterns $P=20$ (decay factor $\gamma_h=0.8$), where highly restrictive similarity thresholds compensate through increased pattern acceptance.

\subsection{Failure Analysis}
\Cref{fig:failure_cases} presents a scenario where Memoir fails despite functioning as designed. In Episode A, the agent must navigate to a bedroom absent from previous episodes. Observation retrieval identifies two distracting bedroom entrances as candidates, while history retrieval surfaces Episodes B (failed) and C (succeeded), both targeting a different bedroom.

\noindent\textbf{Retrieval Limitations.} The retrieved observations prefer incorporating abundant promising candidates as discovered in \Cref{fig:param_search}(a), highlighting both bedroom entrances as semantically relevant but failing to discriminate the critical spatial feature—``nearest to the desk." Retrieved histories similarly cannot distinguish Episodes B and C despite different goals. In Episode B, premature imagination termination after one step limits retrieval to only the nearest entrances, preventing correct target discovery. These failures reveal world model deficiencies in predictive retrieval for both memory types.

\noindent\textbf{Exploration-Exploitation Trade-off.} Episode A fails when both retrieval types converge on the same incorrect location. The agent prioritizes high-similarity histories as discovered in \Cref{fig:param_search}(b), defaulting to exploitation over exploration even when retrieval fails to cover the true goal. It also fails to distinguish task outcomes, treating Episodes B and C equally rather than learning from success. This highlights a new challenge: determining when accumulated experience should be trusted versus when novel alternatives warrant investigation.

\noindent\textbf{Future Work.} These failure modes suggest two potential research directions for advancing imagination-guided memory retrieval. First, \textit{enhanced world modeling capability} through larger-scale pretraining and explicit spatial relationship modeling could address both retrieval inaccuracies in distinguishing spatially distinct targets and premature imagination horizons that affect retrieval scope. Second, \textit{confidence-aware retrieval} to determine when retrieved experience should be trusted, requiring retrieval confidence estimation to dynamically balance exploitation against exploration and serve as a learned filter for memory maintenance to mitigate redundancy. The performance gap between our method (73.46\% SPL) and the oracle retrieval upper bound (93.40\% SPL in \Cref{tab:ablation}) demonstrates significant room for improvement in these directions.

\section{Conclusion}
This work introduces Memoir, a memory-persistent VLN agent employing predictive world modeling for adaptive experience retrieval. Unlike traditional imagine-planning that generates trajectories in isolation, we ground imagination with explicit memory through a language-conditioned world model, Hybrid Viewpoint-Level Memory (HVM) storing observations and behavioral patterns, and an experience-augmented navigation model. Extensive experiments demonstrate 5.4\% SPL improvement on IR2R with 8.3× training speedup and 74\% inference memory reduction, validating that predictive retrieval of both environmental and behavioral memories enables more effective navigation. The oracle retrieval performance (93.4\% SPL) demonstrates the potential of imagination-guided retrieval. Future work should explore enhanced world modeling and confidence-aware exploration mechanisms to narrow this gap, establishing a principled framework connecting predictive simulation with explicit memory for embodied AI.

\bibliographystyle{IEEEtran}
\bibliography{shortstrings,references}

\newpage
\section{Biography Section}

\vspace{-21pt}
\begin{IEEEbiography}[{\includegraphics[width=1in,height=1.25in,clip,keepaspectratio]{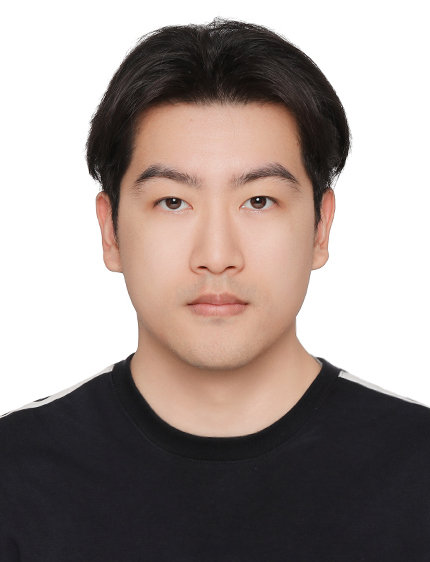}}]{Yunzhe Xu}
received the bachelor's degree in software engineering from Harbin Institute of Technology in 2022. He is currently pursuing the Ph.D. degree in computer science and technology with Shanghai Jiao Tong University. His research interests include embodied navigation system, robotic learning and large language model agents.
\end{IEEEbiography}

\vspace{-21pt}
\begin{IEEEbiography}[{\includegraphics[width=1in,height=1.25in,clip,keepaspectratio]{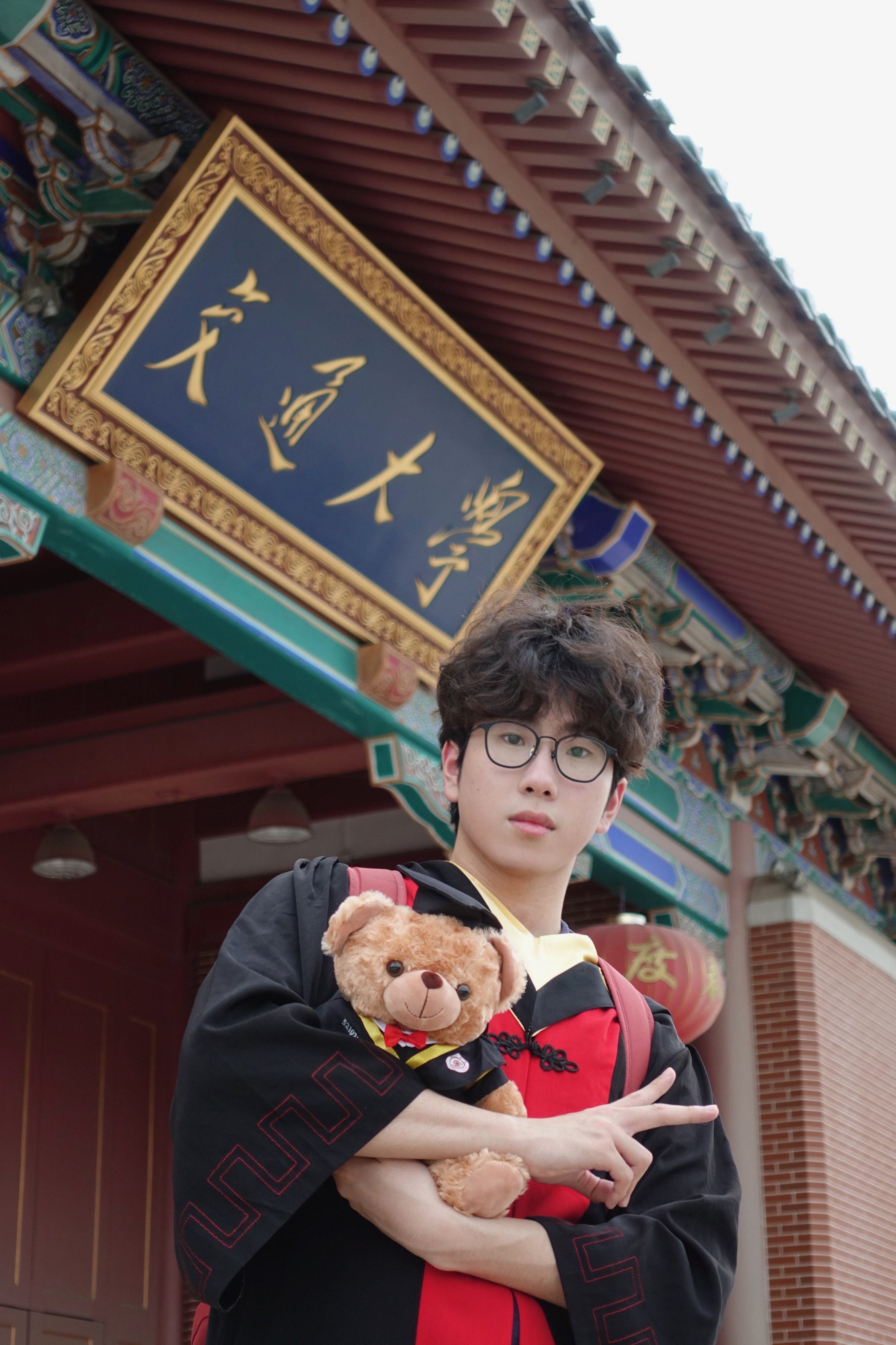}}]{Yiyuan Pan}
received the bachelor's degree in automation from Shanghai Jiao Tong University in 2025. His research focuses on multimodal learning, reinforcement learning and robotic learning. He has published papers in top-tier AI conferences, including AAAI and NeurIPS.
\end{IEEEbiography}

\vspace{-21pt}
\begin{IEEEbiography}[{\includegraphics[width=1in,height=1.25in,clip,keepaspectratio]{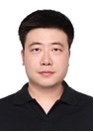}}]{Zhe Liu}
received the Ph.D. degree in control technology and control engineering from Shanghai Jiao Tong University, Shanghai, China, in 2016. From 2017 to 2020, he was a Post-Doctoral Fellow with the Department of Mechanical and Automation Engineering, The Chinese University of Hong Kong, Hong Kong. From 2020 to 2022, he was a Research Associate with the Department of Computer Science and Technology, University of Cambridge, Cambridge, U.K. From 2022 to 2025, he has been an Associate Professor with the AI institute, Shanghai Jiao Tong University, where he is currently an Associate Professor with the Department of Automation. His current research interests include multi-robot cooperation and autonomous driving system.
\end{IEEEbiography}
\vfill

\vfill
\end{document}


\title{Supplementary Material for ``Dream to Recall: Imagination-Guided Experience Retrieval for Memory-Persistent Vision-and-Language Navigation"}


\maketitle

\section{Computational Complexity Analysis}

To validate the efficiency of Memoir, we analyze the time complexity of its three core components: World Model Imagination, Observation Retrieval, and History Retrieval. Let $D$ denote the lookahead horizon (number of imagination steps), $\Lambda$ the maximum spatial density (node count) of a reachable neighborhood shell in the persistent graph, $F$ the feature dimension of state embeddings, and $P$ the number of candidate patterns in the experience memory.

\textbf{World Model Imagination.} 
The world model generates future states in an autoregressive manner. Utilizing a GRU or a Transformer-based architecture with cached attention keys/values, the inference cost per step is dominated by weight projection layers ($\approx F^2$) and remains constant with respect to the sequence length. Consequently, the total time complexity for imagining $D$ steps scales linearly:
\begin{equation}
    T_{img} \approx \mathcal{O}(D \cdot F^2).
\end{equation}

\textbf{Observation Retrieval} 
Standard graph search algorithms on abstract graphs typically suffer from exponential complexity $\mathcal{O}(B^D)$, where $B$ is the branching factor. However, navigation graphs $\mathcal{G}^{(k)}$ in VLN are topological discretizations of 2D physical environments.
In such planar embeddings, the number of nodes at hop-distance $i$, denoted as the shell size $|\mathcal{N}_i(v_t)|$, grows polynomially rather than exponentially.
For short planning horizons ($D \le 5$), we can bound the search space by a spatial density constant $\Lambda = \max_{i \le D} |\mathcal{N}_i(v_t)|$, which represents the maximum number of navigable nodes in the local geometry (e.g., a room or corridor section).
The retrieval algorithm computes compatibility scores for nodes within these shells:
\begin{equation}
    T_{obs} = \sum_{i=1}^{D} |\mathcal{N}_i(v_t)| \cdot F \leq D \cdot \Lambda \cdot F.
\end{equation}
Thus, the complexity scales linearly, $\mathcal{O}(D \cdot \Lambda \cdot F)$. This linear scaling holds because the algorithm exploits the spatial sparsity of the indoor environment, avoiding the combinatorial explosion of unconstrained graph search.

\textbf{History Retrieval.} 
Retrieving relevant experiences involves calculating the similarity between the imagined trajectory sequence (length $D$) and the candidate trajectories stored in memory. This sequence-to-sequence matching requires $\mathcal{O}(D)$ operations per memory candidate:
\begin{equation}
    T_{his} = \mathcal{O}(P \cdot D \cdot F).
\end{equation}
While this term depends on the memory size $P$, it remains linear with respect to the planning horizon $D$.

\textbf{Total Complexity.}
Summing the components for Imagination, Observation Retrieval, and History Retrieval, the overall inference complexity is:
\begin{equation}
    T_{total} = \mathcal{O}\left(D \cdot (F^2 + \Lambda \cdot F + P \cdot F)\right).
\end{equation}
The term $D$ remains a linear factor, confirming the efficiency of the proposed imagination-guided retrieval mechanism.

\section{Implementation Details}

\subsection{World Model Architecture}

Our world model factorizes the latent dynamics through four interconnected components. We first describe the latent state structure, then detail how each component operates.

\subsubsection{Latent State Representation}

The world model maintains a latent state $z_t = [s_t; h_t]$ comprising two components:

\textbf{Deterministic State $h_t \in \mathbb{R}^{d_h}$:} Captures temporal dependencies and sequential patterns through recurrent processing. This component maintains a deterministic trajectory conditioned on language, enabling consistent imagination across time steps.

\textbf{Stochastic State $s_t \in \mathbb{R}^{d_s}$:} Models environmental uncertainty and observation variability through a continuous Gaussian distribution $s_t \sim \mathcal{N}(\mu_t, \sigma_t^2)$. We use $d_s = 96$ for both variants.

The complete state $z_t$ combines deterministic temporal structure with stochastic observation-grounded information.

\subsubsection{Deterministic State Computation}

The deterministic state $h_t$ is computed differently in GRU and Transformer variants:

\textbf{GRU Variant ($d_h = 4000$):} Uses GRU-based recurrence:
\begin{equation}
\begin{aligned}
h_t &= \text{GRU}(s_{t-1}, h_{t-1}): \mathbb{R}^{96} \times \mathbb{R}^{4000} \rightarrow \mathbb{R}^{4000}
\end{aligned}
\end{equation}

\textbf{Transformer Variant ($d_h = 672$):} Uses Transformer decoder with language cross-attention:
\begin{equation}
\begin{aligned}
z_{t-1} &= [s_{t-1}; h_{t-1}] \in \mathbb{R}^{768} \\
h_t &= \text{T5Decoder}(z_{t-1}, \text{cross\_attn}=\hat{\ell}): \mathbb{R}^{768} \rightarrow \mathbb{R}^{768}
\end{aligned}
\end{equation}

\textbf{Initialization:} At $t=1$, both variants initialize $h_0$ from the language [CLS] token:
\begin{equation}
h_0 = \text{MLP}(\hat{\ell}_{\text{[CLS]}}): \mathbb{R}^{768} \rightarrow \mathbb{R}^{d_h}
\end{equation}

\subsubsection{Stochastic State Computation}

Given the deterministic state $h_t$, the stochastic component is computed through two distinct pathways:

\textbf{Transition Model $p(z_t|z_{t-1})$ - Prior Distribution:}

This model predicts the stochastic state purely from temporal dynamics, used during imagination when no observation is available.

For GRU:
\begin{equation}
\begin{aligned}
[\mu_t^{\text{prior}}; \log\sigma_t^{\text{prior}}] &= \text{MLP}(h_t): \mathbb{R}^{4000} \rightarrow \mathbb{R}^{192} \\
s_t^{\text{prior}} &\sim \mathcal{N}(\mu_t^{\text{prior}}, (\sigma_t^{\text{prior}})^2)
\end{aligned}
\end{equation}

For Transformer:
\begin{equation}
\begin{aligned}
[\mu_t^{\text{prior}}; \log\sigma_t^{\text{prior}}] &= \text{MLP}(h_t): \mathbb{R}^{672} \rightarrow \mathbb{R}^{192} \\
s_t^{\text{prior}} &\sim \mathcal{N}(\mu_t^{\text{prior}}, (\sigma_t^{\text{prior}})^2)
\end{aligned}
\end{equation}

\textbf{Inference Model $q(z_t|o_{\leq t}, \ell)$ - Posterior Distribution:}

This model infers the stochastic state from actual observations, grounding the world model in perception during training and state inference.

For GRU:
\begin{equation}
\begin{aligned}
[\mu_t^{\text{post}}; \log\sigma_t^{\text{post}}] &= \text{Linear}(\text{MLP}([x_t; h_t])): \mathbb{R}^{(768 + 4000)} \rightarrow \mathbb{R}^{192} \\
s_t^{\text{post}} &\sim \mathcal{N}(\mu_t^{\text{post}}, (\sigma_t^{\text{post}})^2)
\end{aligned}
\end{equation}

For Transformer:
\begin{equation}
\begin{aligned}
[\mu_t^{\text{post}}; \log\sigma_t^{\text{post}}] &= \text{MLP}([x_t; h_t]): \mathbb{R}^{(768 + 672)} \rightarrow \mathbb{R}^{192} \\
s_t^{\text{post}} &\sim \mathcal{N}(\mu_t^{\text{post}}, (\sigma_t^{\text{post}})^2)
\end{aligned}
\end{equation}

where $x_t \in \mathbb{R}^{768}$ is the observation embedding from the 2-layer observation encoder (hidden dim 768, 12 heads).

\subsubsection{Compatibility Model $p(z_t|o_t)$}

This model enables memory retrieval by measuring state-observation similarity (Eq. 6):
\begin{equation}
\begin{aligned}
e_s &= \psi_s(z_t): \mathbb{R}^{(d_s + d_h)} \rightarrow \mathbb{R}^{256} \\
e_o &= \psi_o(x_t): \mathbb{R}^{768} \rightarrow \mathbb{R}^{256} \\
f(z_t, o_t) &= \frac{1}{\zeta} \frac{e_s^\top e_o}{\|e_s\| \|e_o\|}, \quad p(z_t|o_t) \propto \exp(f(z_t, o_t))
\end{aligned}
\end{equation}

Both projection networks $\psi_s$ and $\psi_o$ are 2-layer MLPs (512→256→256) with ReLU activations. Temperature $\zeta = 0.05$ controls the sharpness of the compatibility distribution.

\subsubsection{Reward Model $p(\gamma_t|z_t)$}

Predicts normalized distance to goal for imagination termination:
\begin{equation}
\hat{\gamma}_t = \text{MLP}(z_t): \mathbb{R}^{(d_s + d_h)} \rightarrow \mathbb{R}
\end{equation}

The 3-layer MLP uses dimensions $(d_s+d_h) \rightarrow 256 \rightarrow 128 \rightarrow 1$ with ReLU activations. Imagination stops when $\hat{\gamma}_{t+i} < \epsilon = 0.15$ or reaches horizon $D = 5$.

\subsection{Navigation Model Architecture}

\subsubsection{Shared Encoders}

\textbf{Text Encoder:} 9-layer BERT-style Transformer ($d_{\text{model}} = 768$, $n_{\text{heads}} = 12$) processes instruction $\ell$ to produce $\hat{\ell} \in \mathbb{R}^{L \times 768}$ for both world model and navigation model.

\textbf{Observation Encoder:} 2-layer Transformer ($d_{\text{model}} = 768$, $n_{\text{heads}} = 12$) shared across components, extracting viewpoint features $x_t \in \mathbb{R}^{768}$ from panoramic observations via average pooling.

\subsubsection{Encoders for Navigation Planning}

The navigation model integrates three information sources:

\textbf{Coarse-Scale Encoder:} 4-layer cross-modal Transformer ($d_{\text{model}} = 768$, $n_{\text{heads}} = 12$) processes retrieved observations with the global topological graph.

\textbf{Fine-Scale Encoder:} 4-layer Transformer ($d_{\text{model}} = 768$, $n_{\text{heads}} = 12$) processes immediate panoramic features $r_t$ for local navigation decisions.

\textbf{Navigation-History Encoder:} 4-layer Transformer ($d_{\text{model}} = 768$, $n_{\text{heads}} = 12$) processes retrieved histories $u_t$ for global navigation decisions.

\subsection{Training Protocol}

\subsubsection{World Model Pretraining}

\begin{itemize}
    \item Dataset: R2R training split + PREVALENT augmented trajectories
    \item Iterations: 5,000, Batch size: 32
    \item Optimizer: AdamW (lr=5e-5, weight decay=0.01)
    \item Temperature: $\zeta = 0.05$, Feature dropout: 0.4
\end{itemize}

\subsubsection{Joint Navigation Training}

\begin{itemize}
    \item \textbf{IR2R:} 10,000 iterations, batch size 8, lr=1e-5 (observation encoder frozen), feature dropout 0.3. Observation retrieval: $\rho_o=0.0$, $W=2$, $\gamma_o=1.0$. History retrieval: $\theta_h=0.6$, $P=10$, $\gamma_h=0.8$. Checkpoint is selected based on SR + SPL.

    \item \textbf{GSA-R2R:} 40,000 iterations, batch size 4, lr=1e-5 (observation encoder frozen), feature dropout 0.4. Observation retrieval: $\rho_o=0.5$, $W=16$, $\gamma_o=1.0$. History retrieval: $\theta_h=0.6$, $P=50$, $\gamma_h=0.7$. Checkpoint is selected based on SR + SPL.
\end{itemize}

\section{Derivations}
\subsection{Basic Variational Bound}

We maximize the joint log-likelihood:

\begin{equation}
\begin{aligned}
\ln \p(o_{1:T},\gamma_{1:T}|\ell) &= \ln \E[\bigg]{\p(s_{1:T}|o_{1:T},\ell)}{\prod_{t=1}^{T} \p(o_t, \gamma_t|s_t)} \\
&= \ln \E[\bigg]{\q(s_{1:T}|o_{1:T},\ell)}{\prod_{t=1}^{T} \frac{\p(o_t,\gamma_t|s_t)\p(s_t|s_{t-1})}{\q(s_t|o_{\leq t},\ell)}} \\
&\geq \E[\bigg]{\q(s_{1:T}|o_{1:T},\ell)}{\ln \prod_{t=1}^{T} \frac{\p(o_t|s_t)\p(\gamma_t|s_t)\p(s_t|s_{t-1})}{\q(s_t|o_{\leq t},\ell)}} \\
&= \E[\bigg]{\q(s_{1:T}|o_{1:T},\ell)}{\sum_{t=1}^{T} \ln \p(o_t|s_t) + \ln \p(\gamma_t|s_t)+\frac{\p(s_t|s_{t-1})}{\q(s_t|o_{\leq t},\ell)}} \\
&= \sum_{t=1}^T \Big( \E{\q(s_t|o_{\leq t},\ell)}{\ln \p(o_t|s_t) + \ln \p(\gamma_t|s_t)} - \E{q(s_{t-1}|o_{\leq t-1},\ell)}{\KL{\q(s_t|o_{\leq t})}{p(s_t|s_{t-1})}} \Big).
\end{aligned}
\label{eq:full_elbo}
\end{equation}

This decomposes into observation reconstruction $\mathcal{J}_{\text{RECOVER}}$, reward prediction $\mathcal{J}_{\text{REWARD}}$, and dynamics regularization $\mathcal{J}_{\text{KL}}$.

\subsection{Contrastive Objective}

We replace expensive reconstruction with contrastive learning:
\begin{equation}
\begin{aligned}
\E{}{\ln \p(o_t|s_t) + \ln \p(\gamma_t|s_t)} &\stackrel{+}{=} \E{}{\ln \p(o_t|s_t) -\ln \p(o_t) + \ln \p(\gamma_t|s_t)} \\[1ex]
&= \E{}{\ln \p(s_t|o_t) -\ln \p(s_t) + \ln \p(\gamma_t|s_t)} \\
&\geq \E[\bigg]{}{\ln \p(s_t|o_t) -\ln \sum_{o'} \p(s_t|o') + \ln \p(\gamma_t|s_t)}. \\
\end{aligned}
\label{eq:full_nce}
\end{equation}

The negative samples $\mathcal{D}$ include observations from different timesteps and episodes within each batch.

\subsection{Multi-Step Bound with Overshooting}

For $d$-step overshooting:

\begin{equation}
\begin{aligned}
\ln \p(o_{1:T},\gamma_{1:T}|\ell) &= \ln \E[\bigg]{\p(s_{1:T}|o_{1:T},\ell)}{\prod_{t=1}^{T} \p(o_t, \gamma_t|s_t)} \\
&\geq \E[\bigg]{\q(s_{1:T}|o_{1:T},\ell)}{\ln \prod_{t=1}^{T} \frac{\p(o_t|s_{t-d+1})\p(\gamma_t|s_{t-d+1})\p(s_t|s_{t-d})}{\q(s_t|o_{\leq t},\ell)}} \\
&= \E[\bigg]{}{\sum_{t=1}^{T} \ln \p(o_t|s_{t-d+1}) + \ln \p(\gamma_t|s_{t-d+1})+ \ln \p(s_t|s_{t-d}) - \ln \q(s_t|o_{\leq t},\ell)} \\
&= \E[\bigg]{}{\sum_{t=1}^{T} \ln \E{\p(s_t|s_{t-d+1})}{\p(o_t|s_t) \p(\gamma_t|s_t)} + \ln \E{\p(s_{t-1}|s_{t-d})}{\p(s_t|s_{t-1})} - \ln \q(s_t|o_{\leq t},\ell)} \\
&\geq \E[\bigg]{}{\sum_{t=1}^{T} \E{\p(s_t|s_{t-d+1})}{\ln \p(o_t|s_t) + \ln \p(\gamma_t|s_t)} + \E{\p(s_{t-1}|s_{t-d})}{\ln \p(s_t|s_{t-1})} - \ln \q(s_t|o_{\leq t},\ell)} \\
&= \sum_{t=1}^T \Big( \E{\p(s_t|s_{t-d+1})\q(s_{t-d+1}|o_{\leq t-d+1},\ell)}{\ln \p(o_t|s_t) + \ln \p(\gamma_t|s_t)} \\
&\hspace{5mm} - \E{\p(s_{t-1}|s_{t-d}) \q(s_{t-d}|o_{\leq t-d},\ell)}{\KL{\q(s_t|o_{\leq t},\ell)}{\p(s_t|s_{t-1})}} \Big) \\
&\geq \sum_{t=1}^T \Big( \E{\p(s_t|s_{t-d+1})\q(s_{t-d+1}|o_{\leq t-d+1},\ell)}{\ln \p(s_t|o_t) -\ln \sum_{o'} \p(s_t|o') + \ln \p(\gamma_t|s_t)} \\
&\hspace{5mm} - \E{\p(s_{t-1}|s_{t-d}) \q(s_{t-d}|o_{\leq t-d},\ell)}{\KL{\q(s_t|o_{\leq t},\ell)}{\p(s_t|s_{t-1})}} \Big). \\
\end{aligned}
\label{eq:full_celbo}
\end{equation}

This encourages accurate long-horizon prediction, critical for imagination-guided retrieval.

\begin{table*}[htbp]
    \renewcommand{\arraystretch}{1.05}
    \setlength{\aboverulesep}{0pt}
    \setlength{\belowrulesep}{0pt}
    \setlength{\tabcolsep}{3.5pt}
    \centering
    \caption{
        The performance of our method on IR2R-CE.
    }
    \resizebox{0.9\textwidth}{!}{
    \begin{threeparttable}
        \begin{tabular}{l c cccccc c cccccc}
            \toprule
            & & \multicolumn{6}{c}{\textbf{Val Seen}} 
            & & \multicolumn{6}{c}{\textbf{Val Unseen}} \\
            \cmidrule{3-8} \cmidrule{10-15}
            Methods & 
            & \textbf{\texttt{TL}}~$\downarrow$
            & \textbf{\texttt{NE}}~$\downarrow$
            & \textbf{\texttt{OS}}~$\uparrow$
            & \textbf{\texttt{nDTW}}~$\uparrow$
            & \textbf{\texttt{SR}}~$\uparrow$
            & \textbf{\texttt{SPL}}~$\uparrow$
            &
            & \textbf{\texttt{TL}}~$\downarrow$
            & \textbf{\texttt{NE}}~$\downarrow$
            & \textbf{\texttt{OS}}~$\uparrow$
            & \textbf{\texttt{nDTW}}~$\uparrow$
            & \textbf{\texttt{SR}}~$\uparrow$
            & \textbf{\texttt{SPL}}~$\uparrow$ \\
            \midrule
            \midrule

            \rowcolor{gray!15}\multicolumn{15}{l}{\emph{Map-based Methods:}}\\
            CMA 
                & & 7.8 & 8.8 & 27  & 42 & 18 & 17
                & & 7.5 & 8.8 & 26 & 44 & 19 & 18 \\
            TourCMA 
                & & 8.0 & 8.2 & 30 & 44 & 20 & 19 
                & & 7.8 & 9.0 & 26 & 42 & 18 & 17 \\
            PoolCMA
                & & 7.2 & 9.1 & 24 & 41 & 17 & 16 
                & & 7.3 & 9.0 & 23 & 42 & 16 & 15 \\
            PoolEndCMA
                & & 7.6 & 8.9 & 27 & 42 & 18 & 17 
                & & 6.9 & 8.7 & 25 & 44 & 18 & 16 \\
            MAP-CMA
                & & 9.4 & 6.4 & 48  & 56 & 39 & 36 
                & & 8.5 & 6.8 & 44 & 54 & 35 & 32 \\
            OVER-NAV
                & & 9.5 & 5.8 & 49 & 59 & 39 & 36 
                & & 8.8 & 6.5 & 45 & 56 & 35 & 33 \\
            
            \midrule

            \rowcolor{gray!15}\multicolumn{15}{l}{\emph{Graph-based Methods:}}\\
            DUET*
                & & 10.5 & 5.0 & 60.9 & 64.7 & 51.8 & 46.6 
                & & 10.6 & \textbf{5.8} & 52.7 & \textbf{58.2} & 44.1 & 37.5 \\
            GR-DUET*
                & & 7.7 & 7.4 & 32.4 & 44.7 & 24.8 & 21.1 
                & & \textbf{8.3} & 7.4 & 30.4 & 44.1 & 23.7 & 18.6 \\
            Memoir (Ours)
                & & 11.1 & 5.2 & 59.3 & 51.1 & 51.1 & 45.7
                & & 12.4 & 6.2 & \textbf{57.1} & 49.2 & \textbf{45.6} & \textbf{39.6} \\
            \bottomrule
        \end{tabular}
        \begin{tablenotes}
            \footnotesize
            \item[*] Results reproduced via the same discrete-to-continuous transfer protocol.
        \end{tablenotes}
    \end{threeparttable}
    }
    \label{tab:ivlnce_results}
\end{table*}

\section{Additional Results} 
\subsection{Evaluation on Continuous Environments}
We further evaluate Memoir on the continuous IR2R-CE benchmark using the standard discrete-to-continuous transfer protocol. As shown in \Cref{tab:ivlnce_results}, Memoir demonstrates robust generalization, achieving state-of-the-art performance (45.6\% SR, 39.6\% SPL) and significantly outperforming the memory-persistent baseline GR-DUET (23.7\% SR). This performance gap highlights Memoir's ability to handle candidate explosion in continuous environments, where the waypoint predictor generates numerous noisy candidates; unlike GR-DUET which indiscriminately incorporates these into memory, Memoir's imagination-guided retrieval effectively filters noise to identify task-relevant waypoints. However, we observe a slight trade-off in path fidelity (lower nDTW compared to single-episode DUET), which stems from the lack of a ground-truth geodesic graph for perfect teacher signal generation during memory updates. This suggests that while retrieval improves goal success, the alignment of retrieved paths with optimal trajectories remains a challenge, warranting future optimization in constructing more accurate persistent topological graphs for continuous spaces.

\begin{figure*}[htbp]
\centering
\includegraphics[scale=0.27]{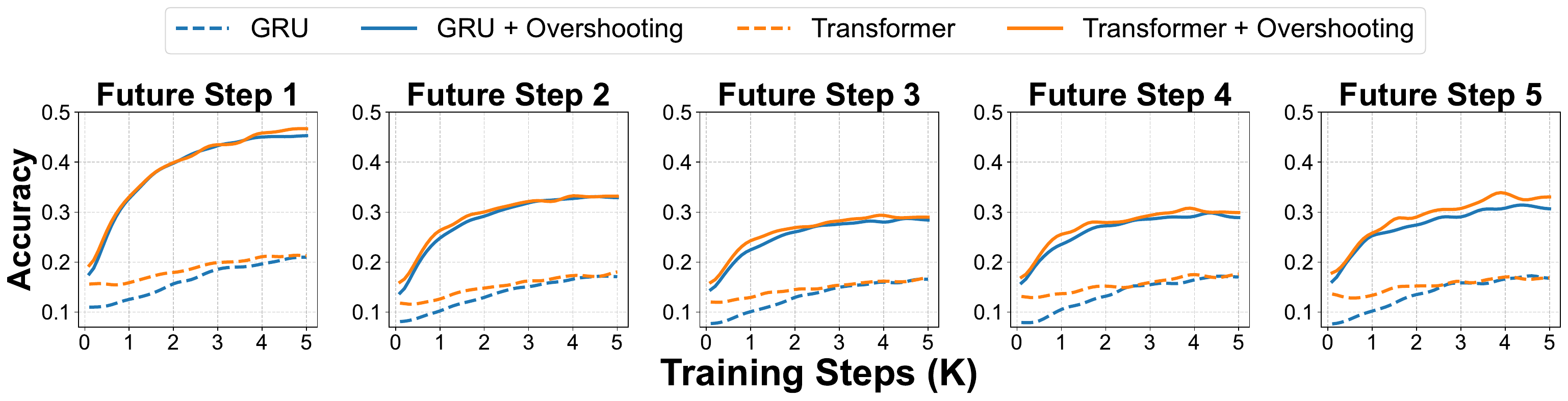}
\caption{Visualization of future observation prediction accuracy across different model variants. Given the current state, models must identify the correct future observation from candidate observations at varying imagination horizon.}
\label{fig:worldmodel}
\end{figure*}

\subsection{World Model Prediction Accuracy}

To further analyze our language-conditioned world model's capability in modeling dynamics, we conduct experiments to evaluate the quality of the compatibility measure between imagined states and observations. For each time step, the world model simulates a trajectory, and the model is tasked with classifying the subsequent observation from a set of distractors. These distractors include observations from other time steps and observations from neighboring nodes along the ground truth trajectory, accumulating with approximately 25 distractors per step for a trajectory.

\Cref{fig:worldmodel} compares different model variants on future observation prediction across 5,000 training steps. We observe that including the overshooting objective significantly improves prediction accuracy over the non-overshooting variants, as validated for both model architectures. This indicates a more robust capability in retrieving episodic memory. Conversely, the model optimized with the single-step variational bound yields poor performance in observation prediction. This stems from the inadequate approximation between the transition model and the inference model, as the compatibility measurement is predominantly conducted between inferred states and observations. By extending the bound to a multi-step, we achieve better observation discrimination, which aids in successful memory retrieval.

\subsection{Detailed Quantitative Results on GSA-R2R}
Due to space constraints in the main manuscript, aggregated performance metrics were presented for the General Scene Adaptation (GSA-R2R) benchmark. In this section, we provide the granular evaluation results broken down by instruction taxonomy and environmental categories.

We report the comprehensive performance comparisons in the following tables:
\begin{itemize}
    \item \textbf{User Instructions (\Cref{tab:gsa_user_results}):} Evaluates adaptation to diverse user personas in residential environments.
    \item \textbf{Scene Instructions (\Cref{tab:gsa_scene_results}):} Assesses performance on descriptions emphasizing scene-based spatial reasoning in non-residential environments.
    \item \textbf{Basic Instructions (\Cref{tab:gsa_basic_results}):} Focuses on standard directional navigational commands fundamental to VLN tasks across residential and non-residential environments.
\end{itemize}

Consistent with the main results, Memoir demonstrates superior performance across these fine-grained splits, validating that the imagination-guided retrieval mechanism offers robust generalization not only across environmental domains but also across varied linguistic styles.

\vfill

\begin{table*}
    \setlength{\tabcolsep}{5pt}
    \renewcommand{\arraystretch}{1.15}
    \setlength{\aboverulesep}{0pt}
    \setlength{\belowrulesep}{0pt}
    \centering
    \caption{
	    Comparison of navigation performance on the GSA-R2R benchmark with user instructions.
	}
	\resizebox{\textwidth}{!}{
    \begin{threeparttable}
		\begin{tabular}{l c cc c cc c cc c cc c cc}
			\toprule
            & &
            \multicolumn{2}{c}{\textbf{Child}}
		   && \multicolumn{2}{c}{\textbf{Keith}} && \multicolumn{2}{c}{\textbf{Moira}} && \multicolumn{2}{c}{\textbf{Rachel}} && \multicolumn{2}{c}{\textbf{Sheldon}}
            \\
			\cmidrule{3-4} \cmidrule{6-7} \cmidrule{9-10} \cmidrule{12-13} \cmidrule{15-16}
			Methods
			&
			& \textbf{\texttt{SR}}~$\uparrow$
			& \textbf{\texttt{SPL}}~$\uparrow$
			&
			& \textbf{\texttt{SR}}~$\uparrow$
			& \textbf{\texttt{SPL}}~$\uparrow$
			&
			& \textbf{\texttt{SR}}~$\uparrow$
			& \textbf{\texttt{SPL}}~$\uparrow$
			&
			& \textbf{\texttt{SR}}~$\uparrow$
			& \textbf{\texttt{SPL}}~$\uparrow$
			&
            & \textbf{\texttt{SR}}~$\uparrow$
			& \textbf{\texttt{SPL}}~$\uparrow$
			\\
			\midrule
			\midrule
                TourHAMT
                & & 14.6 \scriptsize{$\pm$0.2} & 12.0 \scriptsize{$\pm$0.2} & & 15.1 \scriptsize{$\pm$0.2} & 12.3 \scriptsize{$\pm$0.1} & & 13.9 \scriptsize{$\pm$0.1} & 11.3 \scriptsize{$\pm$0.1} & & 15.3 \scriptsize{$\pm$0.1} & 12.5 \scriptsize{$\pm$0.1} & & 14.4 \scriptsize{$\pm$0.1} & 11.8 \scriptsize{$\pm$0.1} \\
                OVER-NAV
                & & 20.9 \scriptsize{$\pm$0.1} & 16.1 \scriptsize{$\pm$0.2} & & 20.5 \scriptsize{$\pm$0.1} & 16.4 \scriptsize{$\pm$0.1} & & 19.5 \scriptsize{$\pm$0.2} & 15.4 \scriptsize{$\pm$0.2} & & 20.6 \scriptsize{$\pm$0.3} & 16.2 \scriptsize{$\pm$0.2} & & 20.5 \scriptsize{$\pm$0.1} & 16.2 \scriptsize{$\pm$0.1}  \\
                \midrule
                DUET
                & & 54.3 & 44.1 & & 56.0 & 46.3 & & 52.3 & 43.3 & & 56.3 & 46.4 & & 54.0 & 44.4 \\
			\quad +MLM
                & & 54.5 \scriptsize{$\pm$0.2} & 44.7 \scriptsize{$\pm$0.2} & & 56.4 \scriptsize{$\pm$0.3} & 46.8 \scriptsize{$\pm$0.3} & & 53.8 \scriptsize{$\pm$0.3} & 43.6 \scriptsize{$\pm$0.4} & & 56.8 \scriptsize{$\pm$0.5} & 46.6 \scriptsize{$\pm$0.6} & & 54.5 \scriptsize{$\pm$0.4} & 44.2 \scriptsize{$\pm$0.3} \\ 
			\quad +MRC
                & & 54.4  \scriptsize{$\pm$0.2} & 44.2 \scriptsize{$\pm$0.1} & & 56.0 \scriptsize{$\pm$0.1} & 46.3 \scriptsize{$\pm$0.1} & & 52.3 \scriptsize{$\pm$0.2} & 43.3 \scriptsize{$\pm$0.1} & & 56.0 \scriptsize{$\pm$0.1} & 46.2 \scriptsize{$\pm$0.2} & & 53.7 \scriptsize{$\pm$0.2} & 44.2 \scriptsize{$\pm$0.4} \\
			\quad +BT
                & & 57.5 \scriptsize{$\pm$0.7} & 54.0 \scriptsize{$\pm$0.9} & & 61.2 \scriptsize{$\pm$0.3} & 57.9 \scriptsize{$\pm$0.1} & & 57.3 \scriptsize{$\pm$0.5} & 54.0 \scriptsize{$\pm$0.6} & & 61.6 \scriptsize{$\pm$0.8} & 58.1 \scriptsize{$\pm$0.7} & & 57.6 \scriptsize{$\pm$0.5} & 54.3 \scriptsize{$\pm$0.5} \\
			\quad +TENT
                & & 54.3 \scriptsize{$\pm$0.2} & 41.7 \scriptsize{$\pm$0.1} & & 55.4 \scriptsize{$\pm$0.2} & 43.8 \scriptsize{$\pm$0.2} & & 51.7 \scriptsize{$\pm$0.2} & 41.0 \scriptsize{$\pm$0.1} & & 55.0 \scriptsize{$\pm$0.2} & 43.2 \scriptsize{$\pm$0.2} & & 53.0 \scriptsize{$\pm$0.2} & 41.9 \scriptsize{$\pm$0.1} \\
		    \quad +SAR
                & & 54.5 \scriptsize{$\pm$0.5} & 41.5 \scriptsize{$\pm$0.4} & & 54.9 \scriptsize{$\pm$0.3} & 43.1 \scriptsize{$\pm$0.2} & & 51.0 \scriptsize{$\pm$0.4} & 40.3 \scriptsize{$\pm$0.6} & & 55.3 \scriptsize{$\pm$0.5} & 43.0 \scriptsize{$\pm$0.6} & & 52.9 \scriptsize{$\pm$0.2} & 41.4 \scriptsize{$\pm$0.4} \\
                \midrule
                \rowcolor{gray!15}\multicolumn{16}{l}{\emph{VLN models pretrained with full navigation graph:}}\\
                GR-DUET
                & & 65.2 \scriptsize{$\pm$0.1} & 59.7 \scriptsize{$\pm$0.1} & & 66.7 \scriptsize{$\pm$0.1} & 62.0 \scriptsize{$\pm$0.1} & & 60.9 \scriptsize{$\pm$0.2} & 56.2 \scriptsize{$\pm$0.2} & & 67.1 \scriptsize{$\pm$0.1} & 62.2 \scriptsize{$\pm$0.1} & & 63.9 \scriptsize{$\pm$0.1} & 58.9 \scriptsize{$\pm$0.1} \\
                GR-DUET*
                & & 64.9 \scriptsize{$\pm$0.5} & 60.5 \scriptsize{$\pm$0.4} & & 65.1 \scriptsize{$\pm$0.3} & 61.4 \scriptsize{$\pm$0.4} & & 60.5 \scriptsize{$\pm$0.3} & 56.6 \scriptsize{$\pm$0.2} & & 65.7 \scriptsize{$\pm$0.5} & 61.7 \scriptsize{$\pm$0.4} & & 63.0 \scriptsize{$\pm$0.4} & 59.0 \scriptsize{$\pm$0.4} \\
                +Memoir
                & & 60.0 \scriptsize{$\pm$0.4} & 49.2 \scriptsize{$\pm$2.3} & & 61.5 \scriptsize{$\pm$0.1} & 52.5 \scriptsize{$\pm$0.1} & & 56.5 \scriptsize{$\pm$0.2} & 47.5 \scriptsize{$\pm$1.2} & & 61.3 \scriptsize{$\pm$0.4} & 52.1 \scriptsize{$\pm$0.1} & & 58.5 \scriptsize{$\pm$0.5} & 49.1 \scriptsize{$\pm$1.0} \\
                +Memoir (Ours)
                & & \bf 66.5 \scriptsize{$\pm$0.5} & \bf 61.3 \scriptsize{$\pm$0.5} & & \bf 68.0 \scriptsize{$\pm$0.1} & \bf 63.6 \scriptsize{$\pm$0.2} & & \bf 62.5 \scriptsize{$\pm$0.3} & \bf 57.5 \scriptsize{$\pm$0.4} & & \bf 68.2 \scriptsize{$\pm$0.1} & \bf 63.6 \scriptsize{$\pm$0.3} & & \bf 65.3 \scriptsize{$\pm$0.1} & \bf 60.4 \scriptsize{$\pm$0.3} \\
			\bottomrule
		\end{tabular}
        \begin{tablenotes}
	    	\footnotesize
	    	\item[*] Results reproduced under aligned experimental conditions (episode ordering, training iterations, batch size, learning rate and dropout rate).
	    \end{tablenotes}
    \end{threeparttable}
    }
	\label{tab:gsa_user_results}
\end{table*}

\begin{table}
    \renewcommand{\arraystretch}{1.15}
    \setlength{\aboverulesep}{0pt}
    \setlength{\belowrulesep}{0pt}
    \centering
    \caption{
	    Comparison of navigation performance on the GSA-R2R benchmark with scene instructions.
	}
	\resizebox{0.6\linewidth}{!}{
    \begin{threeparttable}
		\begin{tabular}{l c ccccc}
			\toprule
            & &
            \multicolumn{5}{c}{\textbf{Test-Non-Residential-Scene}}
            \\
			\cmidrule{3-7}
			Methods
			&
			& \textbf{\texttt{TL}}~$\downarrow$
			& \textbf{\texttt{NE}}~$\downarrow$
			& \textbf{\texttt{SR}}~$\uparrow$
			& \textbf{\texttt{SPL}}~$\uparrow$
            & \textbf{\texttt{nDTW}}~$\uparrow$
			\\
			\midrule
			\midrule
                TourHAMT
                & & 7.3 \scriptsize{$\pm$0.1} & 8.1 \scriptsize{$\pm$0.1} & 9.7 \scriptsize{$\pm$0.1} & 8.0 \scriptsize{$\pm$0.1} & 32.3 \scriptsize{$\pm$0.1} \\
                OVER-NAV
                & & 11.8 \scriptsize{$\pm$0.1} & 7.6 \scriptsize{$\pm$0.2} & 16.7 \scriptsize{$\pm$0.4} & 12.6 \scriptsize{$\pm$0.2} & 34.6 \scriptsize{$\pm$0.3} \\
                \midrule
                DUET
                & & 14.9 & 6.4 & 39.6 & 30.1 & 40.9 \\
			\quad +MLM
                & & 14.3 \scriptsize{$\pm$0.1} & 6.5 \scriptsize{$\pm$0.1} & 39.8 \scriptsize{$\pm$0.1} & 30.5 \scriptsize{$\pm$0.1} & 41.1 \scriptsize{$\pm$0.1} \\
			\quad +MRC
                & & 14.9 \scriptsize{$\pm$0.1} & 6.4 \scriptsize{$\pm$0.1} & 39.7 \scriptsize{$\pm$0.1} & 30.2 \scriptsize{$\pm$0.1} & 40.9 \scriptsize{$\pm$0.1} \\
			\quad +BT
                & & 8.4 \scriptsize{$\pm$0.0} & 6.3 \scriptsize{$\pm$0.2} & 41.2 \scriptsize{$\pm$1.5} & 38.2 \scriptsize{$\pm$1.2} & 51.3 \scriptsize{$\pm$1.2} \\
			\quad +TENT
                & & 16.4 \scriptsize{$\pm$0.1} & 6.3 \scriptsize{$\pm$0.1} & 40.6 \scriptsize{$\pm$0.2} & 28.9 \scriptsize{$\pm$0.2} & 38.9 \scriptsize{$\pm$0.2} \\
		    \quad +SAR
                & & 16.3 \scriptsize{$\pm$0.5} & 6.0 \scriptsize{$\pm$0.2} & 41.4 \scriptsize{$\pm$0.6} & 29.1 \scriptsize{$\pm$0.3} & 39.0 \scriptsize{$\pm$0.3} \\
                \midrule
                \rowcolor{gray!15}\multicolumn{7}{l}{\emph{VLN models pretrained with full navigation graph:}}\\
                GR-DUET
                & & 10.1 \scriptsize{$\pm$0.0} & 5.5 \scriptsize{$\pm$0.0} & 48.1 \scriptsize{$\pm$0.1} & 42.8 \scriptsize{$\pm$0.1} & 53.7 \scriptsize{$\pm$0.1} \\
                GR-DUET*
                & & \bf 9.9 \scriptsize{$\pm$0.3} & 5.5 \scriptsize{$\pm$0.0} & 47.1 \scriptsize{$\pm$0.5} & 42.2 \scriptsize{$\pm$0.8} & 54.1 \scriptsize{$\pm$0.6} \\
                +Memoir
                &
                & 13.5 \scriptsize{$\pm$1.5}
                & 6.2 \scriptsize{$\pm$0.1}
                & 43.3 \scriptsize{$\pm$0.2}
                & 34.1 \scriptsize{$\pm$1.7}
                & 44.2 \scriptsize{$\pm$3.1}
			\\
                +Memoir (Ours)
                &
                & 10.3 \scriptsize{$\pm$0.4}
                & \bf 5.1 \scriptsize{$\pm$0.0}
                & \bf 50.2 \scriptsize{$\pm$0.3}
                & \bf 44.8 \scriptsize{$\pm$0.4}
                & \bf 56.2 \scriptsize{$\pm$0.6}
			\\
			\bottomrule
		\end{tabular}
        \begin{tablenotes}
	    	\footnotesize
	    	\item[*] Results reproduced under aligned experimental conditions.
	    \end{tablenotes}
    \end{threeparttable}
    }
	\label{tab:gsa_scene_results}
\end{table}

\begin{table*}
    \setlength{\tabcolsep}{6pt}
    \renewcommand{\arraystretch}{1.15}
    \setlength{\aboverulesep}{0pt}
    \setlength{\belowrulesep}{0pt}
    \centering
    \caption{
	    Comparison of navigation performance on the GSA-R2R benchmark with basic instructions.
	}
	\resizebox{\textwidth}{!}{
    \begin{threeparttable}
		\begin{tabular}{l c ccccc c ccccc}
			\toprule
            & &
            \multicolumn{5}{c}{\textbf{Test-Residential-Basic}}
		   && \multicolumn{5}{c}{\textbf{Test-Non-Residential-Basic}}
            \\
			\cmidrule{3-7}
			\cmidrule{9-13}
			Methods
			&
			& \textbf{\texttt{TL}}~$\downarrow$
			& \textbf{\texttt{NE}}~$\downarrow$
			& \textbf{\texttt{SR}}~$\uparrow$
			& \textbf{\texttt{SPL}}~$\uparrow$
            & \textbf{\texttt{nDTW}}~$\uparrow$
			&
			& \textbf{\texttt{TL}}~$\downarrow$
			& \textbf{\texttt{NE}}~$\downarrow$
			& \textbf{\texttt{SR}}~$\uparrow$
			& \textbf{\texttt{SPL}}~$\uparrow$
            & \textbf{\texttt{nDTW}}~$\uparrow$
			\\
			\midrule
			\midrule
                TourHAMT
                & & 11.6 \scriptsize{$\pm$0.1} & 7.4 \scriptsize{$\pm$0.1} & 14.9 \scriptsize{$\pm$0.1} & 12.2 \scriptsize{$\pm$0.1} & 34.7 \scriptsize{$\pm$0.1} & & 9.4 \scriptsize{$\pm$0.1} & 7.7 \scriptsize{$\pm$0.1} & 11.0 \scriptsize{$\pm$0.2} & 8.6 \scriptsize{$\pm$0.2} & 32.2 \scriptsize{$\pm$0.1}  \\
                OVER-NAV
                & & 14.1 \scriptsize{$\pm$0.1} & 6.7 \scriptsize{$\pm$0.0} & 22.3 \scriptsize{$\pm$0.3} & 16.8 \scriptsize{$\pm$0.2} & 37.1 \scriptsize{$\pm$0.1} & & 11.4 \scriptsize{$\pm$0.1} & 7.1 \scriptsize{$\pm$0.1} & 16.6 \scriptsize{$\pm$0.2} & 13.0 \scriptsize{$\pm$0.1} & 35.0 \scriptsize{$\pm$0.2} \\
                \midrule
                DUET
                & & 13.1 & 4.2 & 57.7 & 47.0 & 55.6 & & 14.8 & 5.3 & 48.1 & 37.3 & 45.9  \\
			\quad +MLM
                & & 13.1 \scriptsize{$\pm$0.1} & 4.1 \scriptsize{$\pm$0.1} & 57.9 \scriptsize{$\pm$0.2} & 47.3 \scriptsize{$\pm$0.1} & 55.9 \scriptsize{$\pm$0.2} & & 13.1 \scriptsize{$\pm$0.2} & 5.3 \scriptsize{$\pm$0.1} & 48.3 \scriptsize{$\pm$0.5} & 38.8 \scriptsize{$\pm$0.5} & 48.4 \scriptsize{$\pm$0.3} \\
			\quad +MRC
                & & 13.1 \scriptsize{$\pm$0.1} & 4.2 \scriptsize{$\pm$0.1} & 57.7 \scriptsize{$\pm$0.1} & 47.0 \scriptsize{$\pm$0.1} & 55.6 \scriptsize{$\pm$0.1} & & 14.7 \scriptsize{$\pm$0.1} & 5.3 \scriptsize{$\pm$0.1} & 48.1 \scriptsize{$\pm$0.1} & 37.3 \scriptsize{$\pm$0.1} & 45.9 \scriptsize{$\pm$0.1} \\
			\quad +BT
                & & 8.0 \scriptsize{$\pm$0.1} & 3.8 \scriptsize{$\pm$0.1} & 61.3 \scriptsize{$\pm$0.6} & 57.7 \scriptsize{$\pm$0.3}& 70.1 \scriptsize{$\pm$0.5}& & 7.9 \scriptsize{$\pm$0.0} & 5.2 \scriptsize{$\pm$0.1} & 49.5 \scriptsize{$\pm$0.8} & 46.0 \scriptsize{$\pm$0.8} & 59.4 \scriptsize{$\pm$0.9}  \\
			\quad +TENT
                & & 14.6 \scriptsize{$\pm$0.0} & 4.2 \scriptsize{$\pm$0.0} & 57.2 \scriptsize{$\pm$0.4} & 44.2 \scriptsize{$\pm$0.4} & 52.9 \scriptsize{$\pm$0.1} & & 16.2 \scriptsize{$\pm$0.1} & 5.4 \scriptsize{$\pm$0.1} & 46.5 \scriptsize{$\pm$0.4} & 33.7 \scriptsize{$\pm$0.2} & 42.6 \scriptsize{$\pm$0.3} \\
		    \quad +SAR
                & & 13.8 \scriptsize{$\pm$0.8} & 4.0 \scriptsize{$\pm$0.1} & 57.6 \scriptsize{$\pm$0.2} & 44.6 \scriptsize{$\pm$0.2} & 53.0 \scriptsize{$\pm$0.2} & & 16.5 \scriptsize{$\pm$0.0} & 5.4 \scriptsize{$\pm$0.0} & 44.6 \scriptsize{$\pm$1.5} & 31.5 \scriptsize{$\pm$1.6} & 40.6 \scriptsize{$\pm$1.3} \\
                \midrule
                \rowcolor{gray!15}\multicolumn{13}{l}{\emph{VLN models pretrained with full navigation graph:}}\\
                GR-DUET
                & & 9.4 \scriptsize{$\pm$0.0} & 3.1 \scriptsize{$\pm$0.0} & 69.3 \scriptsize{$\pm$0.2} & 64.3 \scriptsize{$\pm$0.1} &  71.4 \scriptsize{$\pm$0.1} & & 8.9 \scriptsize{$\pm$0.0} & 4.4 \scriptsize{$\pm$0.0} & 56.6 \scriptsize{$\pm$0.1} & 51.5 \scriptsize{$\pm$0.1} & 61.0 \scriptsize{$\pm$0.1} \\
                GR-DUET*
                & & \bf 8.6 \scriptsize{$\pm$0.2} & 3.2 \scriptsize{$\pm$0.1} & 67.6 \scriptsize{$\pm$0.5} & 63.6 \scriptsize{$\pm$0.6} &  71.9 \scriptsize{$\pm$0.5} & & \bf 8.7 \scriptsize{$\pm$0.4} & 4.4 \scriptsize{$\pm$0.0} & 55.3 \scriptsize{$\pm$0.2} & 50.4 \scriptsize{$\pm$0.3} & 60.8 \scriptsize{$\pm$0.4} \\
                +Memoir
                &
                & 11.7 \scriptsize{$\pm$0.1}
                & 3.7 \scriptsize{$\pm$0.0}
                & 63.0 \scriptsize{$\pm$0.3}
                & 52.9 \scriptsize{$\pm$0.3}
                & 61.0 \scriptsize{$\pm$0.5}
                &
                & 12.6 \scriptsize{$\pm$0.3}
                & 4.9 \scriptsize{$\pm$0.1}
                & 51.6 \scriptsize{$\pm$0.8}
                & 40.8 \scriptsize{$\pm$0.1}
                & 49.6 \scriptsize{$\pm$0.0} \\
                +Memoir (Ours)
                &
                & 9.3 \scriptsize{$\pm$0.0}
                & \bf 3.0 \scriptsize{$\pm$0.0}
                & \bf 69.8 \scriptsize{$\pm$0.2}
                & \bf 64.9 \scriptsize{$\pm$0.4}
                & \bf 73.3 \scriptsize{$\pm$0.2}
                &
                & 9.3 \scriptsize{$\pm$0.2}
                & \bf 4.2 \scriptsize{$\pm$0.0}
                & \bf 57.7 \scriptsize{$\pm$0.1}
                & \bf 52.0 \scriptsize{$\pm$0.1}
                & \bf 61.9 \scriptsize{$\pm$0.4}
			\\
			\bottomrule
		\end{tabular}
        \begin{tablenotes}
	    	\footnotesize
	    	\item[*] Results reproduced under aligned experimental conditions (episode ordering, training iterations, batch size, learning rate and dropout rate).
	    \end{tablenotes}
    \end{threeparttable}
    }
	\label{tab:gsa_basic_results}
\end{table*}